\documentclass[11pt]{article}

\usepackage[final]{acl}

\usepackage{times}
\usepackage{latexsym}

\usepackage[T1]{fontenc}

\usepackage[utf8]{inputenc}

\usepackage{microtype}

\usepackage{inconsolata}

\usepackage{graphicx}
\usepackage{amsmath}
\usepackage{amssymb}
\usepackage{booktabs}
\usepackage{adjustbox}
\usepackage{multirow}
\usepackage[table]{xcolor}

\title{Towards Detecting AI-Assisted Responses in Online Surveys}

\renewcommand{\thefootnote}{\fnsymbol{footnote}}

\author{
  \textbf{Qizhou Wang\textsuperscript{1}}\thanks{~Equal contribution.}\thanks{~Corresponding author.}
  \quad
  \textbf{Bogdan Mamaev\textsuperscript{2}}\footnotemark[1]
  \quad
  \textbf{Christopher Leckie\textsuperscript{1}}
\\[2pt]
  \textsuperscript{1}The University of Melbourne, Australia
  \quad
  \textsuperscript{2}Deakin University, Australia
\\[2pt]
  \texttt{\{mike.wang, caleckie\}@unimelb.edu.au}
  \quad
  \texttt{bogdan.mamaev@deakin.edu.au}
}

\begin{document}
\maketitle
\renewcommand{\thefootnote}{\arabic{footnote}}
\setcounter{footnote}{0}

\begin{abstract}

The use of LLMs to complete online surveys impacts the validity of survey-based research, but detecting such usage remains under-explored. We introduce an initial benchmark dataset, namely ASURRE, for AI-assisted survey participation to capture usage strategies ranging from full generation and revision to persona-grounded agentic completion. Controlled by these strategies, LLM-assisted survey responses are generated using multiple LLMs on three real-world surveys in different disciplines, paired with genuine human responses. Our evaluation of existing machine-generated text (MGT) detectors shows that naive AI usage is readily detectable, whereas persona-grounded agents that mimic entire respondents push detector performance toward chance. We further show that agentic completion cannot fully replicate respondent-level behaviour and leaves distinctive behavioural traces. While individual cues can be circumvented by targeted prompting, a simple few-shot, training-free aggregator over these cues improves mean AUROC by +0.14 over the best existing detector across agentic settings. Our project is available on GitHub. \footnote{\url{https://github.com/mike-qz-wang/ASURRE}}.

\end{abstract}

\section{Introduction}

Online surveys are one of the main tools of social science research used to measure attitudes, behaviours, and social phenomena \citep{groves2011survey}. However, the growing use of generative AI tools raises a question: can survey responses still be trusted as genuine human input? Under incentives such as reducing effort, maximising compensation, or manipulating outcomes, participants may rely on AI models, which threatens validity assumptions for most survey instruments \citep{krosnick1991response, meade2012identifying, huang2012detecting}. For instance, \citet{zhang2025generative} find that 34\% of participants on a major crowdsourcing platform reported using LLMs to answer open-ended survey questions, with AI-generated responses systematically more homogeneous than human ones. Beyond self-reported usage, AI-assisted submissions have been identified as a documented validity risk in survey data collection more broadly \citep{veselovsky2023artificial, kennedy2020shape}. This risk is increased by evidence that persona-grounded LLMs can reproduce human survey responses that can pass as genuine \citep{argyle2023one, park2024generative}. However, such studies do not discuss detection, nor is data with raw responses easily accessible to run such analyses. Thus, detecting AI usage in online surveys is increasingly important. 

To study this underexplored problem, we construct a benchmark dataset, namely AI-assisted SURvey REsponses (ASURRE). It captures revision, full generation, and persona-grounded agentic completion as three usage strategies across three real-world surveys (App.~\ref{sec:datasets}) and multiple open and proprietary LLMs, covering both question-level and response-level usage. We evaluate existing MGT detection models \citep{wu2025survey, yang2023survey} from two families: training-free approaches exploiting token-level probability statistics \citep{mitchell2023detectgpt, bao2024fastdetectgpt, solaiman2019release, gehrmann2019gltr, hans2024binoculars} and supervised classifiers trained on labelled MGT corpora \citep{hu2023radar}. However, these detectors are intended to be used on long-form text, and not on short, often fragmented responses that are spread across multiple fields in a survey. They are not able to capture respondent behaviour from response to response.

Our key finding is that AI use can be identified not only in individual answers, but also in behavioural patterns across a respondent's full set of answers. Detection can therefore extend beyond single answers to respondent-level analysis. In this regard, we make three contributions: (1) a benchmark dataset for AI-assisted survey participation, covering three usage patterns across three surveys and several LLMs; (2) a systematic evaluation showing that detection difficulty varies by usage pattern — full generation is detectable, revision is partially detectable, and agentic completion is identified near chance — confirming that the most realistic threat is the least detectable; and (3) SPABD, a simple aggregator over four behavioural cues computed across a respondent's answers, which requires no training and only a few reference responses, and improves mean AUROC from 0.61 to 0.75 in agentic settings.


\section{Detecting AI Survey Completion}
\label{sec:gensim}

\noindent\textbf{Problem Definition.}
\label{sec:problem}
For each respondent $r$, we observe a response set $R_r = \{(q_k, a_{r,k})\}_{k}$ of answered questions $q_k$ and corresponding text $a_{r,k}$. Each $R_r$ is produced either by a human or by one of the three AI usage strategies introduced below: \emph{revision}, \emph{full generation}, and \emph{agentic completion}. We distinguish two detection granularities. A \emph{per-question} detector scores each answer in isolation and returns $f(q_k, a_{r,k}) \in [0,1]$. A \emph{per-respondent} detector aggregates the respondent's full answer set into a single score $f(R_r) \in [0,1]$, so it can exploit cross-answer signals that no single answer reveals, such as length variance and engagement levels. Existing MGT detectors tend to focus on individual answers without considering the full answer set.

\subsection{Dataset Creation}
\label{sec:dataset_creation}
Our dataset pairs real human responses from three publicly available surveys with AI submissions across three usage strategies and multiple LLMs. Full details on datasets, simulation pipeline, and persona portraits are in App.~\ref{sec:simulation}. \smallskip

\noindent \textbf{Question-Level Completion.} Two usages operate independently per question. \emph{Revision} rewrites a respondent's own answer for stylistic polish, preserving meaning, opinions, and concrete details. \emph{Full Generation} produces a plausible but ungrounded answer from the question text alone. \smallskip

\noindent \textbf{Agentic Completion.} A persona-grounded LLM simulator mimics a specific real respondent across the entire survey \citep{argyle2023one, park2024generative}. From the respondent's answer set, an LLM first produces a structured persona portrait covering writing style, affect, English level, seriousness, themes, and data-derived length habits. A fresh simulator then fills the survey question by question, conditioned only on the portrait and its own prior answers, and autonomously decides which questions to answer or skip. 

A final humanisation pass suppresses AI stylistic markers. We evaluate four prompting modes: natural, controlled, length-restrictive, and combined. Specifically, natural means that the agent follows only the persona portrait and completes the survey with full autonomy. The other three are humanisation variants targeting specific behavioural cues. Controlled adds a skip policy so the agent can skip questions based on the persona portrait, while length-restrictive constrains verbosity. Combined applies both. \smallskip

\noindent \textbf{Datasets and Models.} We use three publicly available surveys, all collected before LLMs became widely available (pre-2020), so that every original response is verifiably human-written: \textbf{phd2019} (\textit{Nature} PhD Students Survey 2019), \textbf{springer2017} (Springer Nature Social Media Survey 2017), and \textbf{osmi\_mh} (OSMI Mental Health in Tech Survey 2017--18). Tab.~\ref{tab:dataset_structure} reports the per-usage train/test composition. Revision and Full Generation use three augmenter LLMs: \texttt{gpt-oss-120b} (OSS120; \citealp{openai2025gptoss}), \texttt{Qwen3.5-35B-A3B} (Qwen3.5; \citealp{yang2025qwen3}), and Gemini-3-Flash-Preview (Gemini3; \citealp{google2025gemini3}). Agentic Completion uses a Claude Code agent as the persona simulator, with Sonnet 4.6 \citep{anthropic2025sonnet} as its primary backbone and OSS120 and Gemini3 reserved for the cross-LLM ablation (Sec.~\ref{sec:ablations}).

\begin{table}[t]
\centering
\begin{adjustbox}{max width=\columnwidth}
\begin{tabular}{lccc}
\toprule
            & Train H+AI                 & Test H+AI                & Agent Test H+AI \\
Dataset     & $\times$ 3 LLMs            & $\times$ 3 LLMs          & $\times$ 4 modes \\
\midrule
phd2019      & (2725+2725)\,$\times$\,3 & (682+681)\,$\times$\,3  & (504+504)\,$\times$\,4 \\
springer2017 & (638+638)\,$\times$\,3   & (160+160)\,$\times$\,3  & (90+90)\,$\times$\,4   \\
osmi\_mh     & (427+427)\,$\times$\,3   & (107+107)\,$\times$\,3  & (107+107)\,$\times$\,4 \\
\bottomrule
\end{tabular}
\end{adjustbox}
\caption{Dataset train/test composition. Revision and Full Generation share train/test splits across three augmenter LLMs; Agentic Completion is test-only across four prompting modes.}
\label{tab:dataset_structure}
\end{table}

\subsection{Analysing Human vs.\ AI Completion}
\label{sec:analysis}

We analyse human--AI divergence through four respondent-level behavioural cues (Tab.~\ref{tab:cues}). The cues require only a sentence encoder (\texttt{all-MiniLM-L6-v2}; \citealp{reimers2019sbert, wang2020minilm}) and no reference scoring LLM, training labels, or population calibration; each captures a behavioural property of LLM generation that humans need not follow.

\begin{table}[t]
\centering
\begin{adjustbox}{scale=0.8}
\begin{tabular}{lp{5cm}c}
\toprule
Cue & Definition & AI dir. \\
\midrule
\texttt{min\_len}     & shortest answered field (words)            & $\uparrow$ \\
\texttt{length\_cv}   & std/mean of answer lengths                  & $\downarrow$ \\
\texttt{mean\_qa\_sim} & mean cos.\ sim.\ between answer and its question & $\uparrow$ \\
\texttt{n\_answered}  & number of answered fields                   & $\uparrow$\,(nat.) \\
\bottomrule
\end{tabular}
\end{adjustbox}
\caption{Four respondent-level behavioural cues. ``AI dir.'' is the sign of the AI--human distributional shift.}
\label{tab:cues}
\end{table}

\emph{Full Generation} is the easy case for existing MGT detectors (Sec.~\ref{sec:detector_bench}, Tab.~\ref{tab:main_results}) because the AI signal lives at the token level. Behavioural cues still show up here, most clearly the verbosity gap: AI per-field answers are roughly twice as long as human ones. \emph{Revision} preserves human length, content, and coverage by construction, leaving only style cues for detection. 

\emph{Agentic Completion} is where behavioural cues become necessary. Persona completions diverge from humans on all four cues: they over-answer optional fields (\emph{n\_answered}; OSMI 10 vs 2), produce uniform-length responses (\emph{length\_cv}; PhD 0.33 vs 0.67), respect a minimum-length floor that humans freely violate (\emph{min\_len}; PhD 6 vs 2 words), and answer slightly more on-topic (\emph{mean\_qa\_sim}; small but consistent shift). At the field level, however, the agentic outputs are visually indistinguishable from one another and often from human prose (qualitative examples in App.~\ref{sec:qualitative}), motivating the respondent-level aggregation in Sec.~\ref{sec:spabd}. The pattern holds across all three surveys and four prompting modes (App.~\ref{sec:appendix_b}, Tab.~\ref{tab:appendix-b1} and Figs.~\ref{fig:appendix_b1}, \ref{fig:appendix_b2}).

\subsection{Aggregating Respondent-Level Behavioural Cues}
\label{sec:spabd}

We combine the four cues into SPABD (Survey-Pool Anchored Behavioural Detection), a few-shot, training-free detector. Given $K$ known-human reference responses, for each cue $c$ with direction prior $\mathrm{sgn}(c)$ (Tab.~\ref{tab:cues}), we compute a robust location $\mu_c = \mathrm{median}_{r \in R}\, c(r)$ and scale $\sigma_c = \mathrm{median}_{r \in R}\, |c(r) - \mu_c|$. The aggregate score for test respondent $r$ is
\[
\mathrm{Score}(r) \;=\; \max_c \left( \mathrm{sgn}(c) \cdot \frac{c(r) - \mu_c}{\sigma_c} \right),
\]
a one-tail, prior-corrected max-aggregator. We rank by $\mathrm{Score}$ and flag top-tail respondents as AI. The design has three properties: \emph{(i)} median/MAD on $K$ known humans provides a contamination-free reference distribution; \emph{(ii)} sign priors derived from LLM-generation properties preserve directional information through one-tail aggregation; \emph{(iii)} max-aggregation ensures no single targeted attack on one cue defeats the score (Sec.~\ref{sec:cue_eval}).

\section{Experiments}

We evaluate whether AI-assisted survey participation can be detected under increasingly realistic usage patterns. We first benchmark existing MGT detectors, then evaluate SPABD against the strongest existing detector in agentic-completion settings. We focus on zero-shot and training-free methods because labelled AI-assisted survey data is unlikely to be available for new surveys, and per-survey fine-tuning is often impractical.

We benchmark six existing detectors zero-shot: a prompt-based classifier using \texttt{gpt-oss-120b} \citep{openai2025gptoss}; token-probability methods including DetectGPT \citep{mitchell2023detectgpt}, Fast-DetectGPT \citep{bao2024fastdetectgpt}, and per-token log-likelihood \citep{solaiman2019release, gehrmann2019gltr}; Binoculars \citep{hans2024binoculars}; and the supervised detector RADAR \citep{hu2023radar}. These baselines are mostly training-free, reflecting the lack of survey-labelled AI usage data, while RADAR tests whether general MGT training transfers to survey responses. Since they score individual texts, we average answer-level scores to obtain respondent-level scores. Details are in App.~\ref{sec:detectors}. We report \textbf{AUROC}, the conventional metric in AI-text detection  \citep{hans2024binoculars, dugan2024raid}.
\begin{figure}[!t]
\centering
\includegraphics[width=\columnwidth]{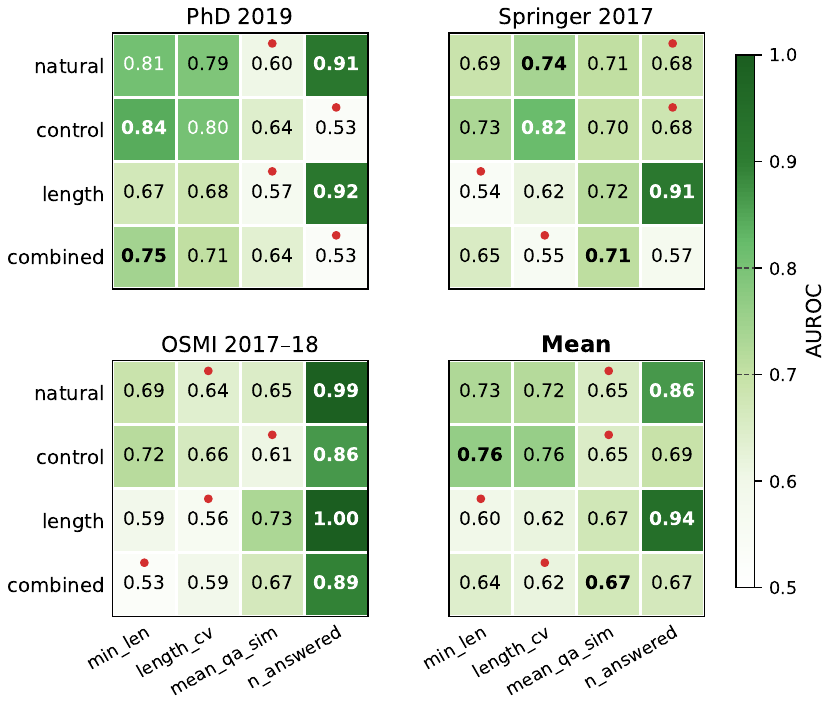}
\caption{Cue AUROC across four prompting modes (natural, controlled, length-restrictive, combined) and three surveys. Rows: prompting modes; columns: cues.}
\label{fig:orthogonality}
\vspace{-0.1cm}
\end{figure}

\subsection{Zero-Shot Performance}
\label{sec:detector_bench}
Tab.~\ref{tab:main_results} reports test-split AUROC. Existing detectors perform best under Full Generation, where top AUROCs reach 0.79--0.93, but degrade under Revision (0.55--0.74) and further under Agentic Completion, where most scores remain close to chance (0.50--0.65). Log-Likelihood and Fast-DetectGPT obtain the strongest row averages, but their advantage is driven mainly by Full Generation and does not transfer to Agentic Completion once responses are conditioned on persona context and prior answers. RADAR performs below chance on phd2019 (0.28) and springer2017 (0.39), suggesting a domain mismatch between its training distribution and short open-ended survey responses. The prompt-based Zero-Shot classifier is also  near chance ($\sim$0.50--0.53). Per-augmenter results are in App.~\ref{sec:full_results}.
\begin{table}[t]
\centering
\begin{adjustbox}{max width=1.0\columnwidth}
\begin{tabular}{llcc *{4}{c}}
\toprule
& & Rev. & Gen. & \multicolumn{4}{c}{Agentic Fill} \\
\cmidrule(lr){3-3} \cmidrule(lr){4-4} \cmidrule(lr){5-8}
D.set & Method & avg & avg & \textsc{nat} & \textsc{ctrl} & \textsc{len} & \textsc{cmb} \\
\midrule
\multirow{6}{*}{phd}
 & Zero-shot   & 0.63 & 0.45 & 0.43 & 0.45 & 0.44 & 0.46 \\
 & Fast-DGPT   & 0.64 & 0.87 & 0.50 & 0.55 & 0.54 & 0.53 \\
 & DetectGPT   & 0.57 & 0.83 & 0.61 & 0.59 & 0.56 & 0.57 \\
 & Log-Lik     & 0.66 & 0.90 & 0.56 & 0.57 & 0.57 & 0.58 \\
 & Binoculars  & 0.57 & 0.69 & 0.63 & 0.59 & 0.63 & 0.61 \\
 & RADAR       & 0.29 & 0.25 & 0.28 & 0.34 & 0.24 & 0.23 \\
\midrule
\multirow{6}{*}{sp}
 & Zero-shot   & 0.60 & 0.45 & 0.47 & 0.41 & 0.51 & 0.52 \\
 & Fast-DGPT   & 0.74 & 0.88 & 0.58 & 0.62 & 0.52 & 0.53 \\
 & DetectGPT   & 0.64 & 0.87 & 0.61 & 0.64 & 0.51 & 0.56 \\
 & Log-Lik     & 0.64 & 0.93 & 0.62 & 0.68 & 0.52 & 0.56 \\
 & Binoculars  & 0.44 & 0.68 & 0.56 & 0.55 & 0.56 & 0.59 \\
 & RADAR       & 0.34 & 0.39 & 0.42 & 0.51 & 0.35 & 0.31 \\
\midrule
\multirow{6}{*}{osmi}
 & Zero-shot   & 0.63 & 0.49 & 0.45 & 0.44 & 0.45 & 0.45 \\
 & Fast-DGPT   & 0.64 & 0.79 & 0.46 & 0.48 & 0.44 & 0.46 \\
 & DetectGPT   & 0.49 & 0.81 & 0.60 & 0.65 & 0.55 & 0.56 \\
 & Log-Lik     & 0.62 & 0.78 & 0.45 & 0.48 & 0.40 & 0.44 \\
 & Binoculars  & 0.55 & 0.68 & 0.53 & 0.51 & 0.55 &  0.52 \\
 & RADAR       & 0.40 & 0.62 & 0.56 & 0.59 & 0.50 & 0.53 \\
\bottomrule
\end{tabular}
\end{adjustbox}
\caption{Test AUROC of the baselines. \emph{Rev.}\ and \emph{Gen.}\ are averages over the three augmenter LLMs; Agentic Fill columns are per-mode (\textsc{nat}/\textsc{ctrl}/\textsc{len}/\textsc{cmb} = natural / controlled / length-restrictive / combined).}
\label{tab:main_results}
\end{table}

\subsection{Behavioural Cue Evaluation}
\label{sec:cue_eval}

Fig.~\ref{fig:orthogonality} shows an oracle cue analysis that measures the discriminability of each respondent-level cue and demonstrates that they provide additional signal across datasets and prompting modes. Not a single cue dominates consistently. For instance, coverage becomes weaker when agents are prompted to skip survey fields, while length weakens under length-restrictive prompting. Even so, other cues often remain informative, indicating that targeted prompting can reduce individual behavioural artefacts. However, it does not eliminate all respondent-level signals as multiple cues aggregate.

\subsection{SPABD Performance}
\label{sec:spabd_perf}

Tab.~\ref{tab:spabd_perf} compares SPABD ($K{=}5$, one-tail prior max aggregation) with the best existing MGT detector in each of the 12 agentic-completion settings. SPABD improves mean AUROC from 0.61 to 0.75, showing that respondent-level behavioural cues provide complementary signal beyond per-answer text detection. It outperforms the best existing detector in 11 of 12 settings, with the only exception being the combined adversarial mode on Springer (0.56 $\pm$ 0.10), where multiple cues are simultaneously weakened. Performance remains variable across surveys and prompting modes, suggesting that SPABD should be viewed as a lightweight reference detector rather than a complete solution. 

Tab.~\ref{tab:deployment-metrics} reports the aggregated operating-point performance across the same settings. This is more indicative of detection performance at a fixed threshold. SPABD exhibits advantageous performance under tight false-positive tolerance. Specifically, at FPR=5\%, it detects 27.7\% of AI submissions versus 12.0\% for the best existing detector, and at FPR=1\%, 15.5\% versus 4.5\%. On the shortest survey, springer2017, the best existing detectors remain stronger at low FPR. However, it is also worth noting that the tolerance is a hyperparameter chosen in practice, and SPABD performs better overall in both types of metrics  by a larger margin. Metric definitions are in App.~\ref{sec:deployment}.

\begin{table}[t]
\centering
\begin{adjustbox}{max width=\columnwidth}
\begin{tabular}{l *{4}{c} c}
\toprule
Method & Nat & Ctrl & Len & Cmb & Mean \\
\midrule
\multicolumn{6}{c}{\textit{phd2019}} \\
\cmidrule(lr){1-6}
Best exist & 0.63 & 0.59 & 0.63 & 0.61 & 0.61 \\
SPABD      & \textbf{0.82}\,$\pm$.05 & \textbf{0.79}\,$\pm$.04 & \textbf{0.72}\,$\pm$.07 & \textbf{0.68}\,$\pm$.03 & \textbf{0.75} \\
\midrule
\multicolumn{6}{c}{\textit{springer2017}} \\
\cmidrule(lr){1-6}
Best exist & 0.62 & 0.68 & 0.56 & 0.59 & 0.61 \\
SPABD      & \textbf{0.75}\,$\pm$.06 & \textbf{0.72}\,$\pm$.11 & \textbf{0.72}\,$\pm$.11 & 0.56\,$\pm$.10 & \textbf{0.69} \\
\midrule
\multicolumn{6}{c}{\textit{osmi\_mh}} \\
\cmidrule(lr){1-6}
Best exist & 0.60 & 0.65 & 0.55 & 0.56 & 0.59 \\
SPABD      & \textbf{0.89}\,$\pm$.12 & \textbf{0.76}\,$\pm$.07 & \textbf{0.88}\,$\pm$.16 & \textbf{0.73}\,$\pm$.12 & \textbf{0.82} \\
\midrule
\multicolumn{6}{c}{\textit{Mean}} \\
\cmidrule(lr){1-6}
Best exist & 0.62 & 0.64 & 0.58 & 0.59 & 0.61 \\
SPABD      & \textbf{0.82} & \textbf{0.76} & \textbf{0.77} & \textbf{0.66} & \textbf{0.75} \\
\bottomrule
\end{tabular}
\end{adjustbox}
\caption{SPABD ($K{=}5$, 1-tail prior max) vs best per-cell existing detector AUROC; SPABD = mean $\pm$ std over 50 seeds. Bold = best per cell.}
\label{tab:spabd_perf}
\vspace{-0.3cm}
\end{table}

\begin{table}[t]
\centering
\begin{adjustbox}{max width=\columnwidth}
\begin{tabular}{ll cccc}
\toprule
D.set & Method & AUROC & \textsc{tpr@}1\% & \textsc{tpr@}5\% & \textsc{tpr@}10\% \\
\midrule
\multirow{2}{*}{phd}  & Best exist & 0.613 & 0.018 & 0.085 & 0.175 \\
                      & SPABD      & \textbf{0.753} & \textbf{0.157} & \textbf{0.279} & \textbf{0.317} \\
\midrule
\multirow{2}{*}{sp}   & Best exist & 0.613 & \textbf{0.064} & \textbf{0.153} & \textbf{0.230} \\
                      & SPABD      & \textbf{0.687} & 0.008 & 0.105 & 0.195 \\
\midrule
\multirow{2}{*}{osmi} & Best exist & 0.591 & 0.052 & 0.124 & 0.218 \\
                      & SPABD      & \textbf{0.813} & \textbf{0.301} & \textbf{0.446} & \textbf{0.497} \\
\midrule
\multirow{2}{*}{Mean} & Best exist & 0.606 & 0.045 & 0.120 & 0.207 \\
                      & SPABD      & \textbf{0.751} & \textbf{0.155} & \textbf{0.277} & \textbf{0.336} \\
\bottomrule
\end{tabular}
\end{adjustbox}
\caption{Operating-point metrics on the 12 agent\_fill settings with SPABD at $K{=}5$. \emph{Best exist} is the best existing detector per setting. \textbf{TPR@$x$\%} is the AI detection rate at $x\%$ FPR.}
\label{tab:deployment-metrics}
\end{table}

\subsection{Ablation Studies}
\label{sec:ablations}

We ablate five design choices in SPABD: reference-set size, persona-simulator backbone, aggregator, the number of open-ended fields, and the reference sample quality. \smallskip 

\noindent\textbf{Scaling with $K$.} SPABD's reference set is the only supervision it consumes; we ablate its size. Performance scales smoothly across $K \in \{5,10,20,50,100\}$ (mean AUROC 0.75 $\to$ 0.79), with $K{=}5$ capturing most of the achievable gain and the seed-level std shrinking from 0.09 to 0.02 (Tab.~\ref{tab:k_scan}, App.~\ref{sec:ablation_tables}). Five known-human responses (which a survey practitioner can supply from a pre-LLM-era split or small curated set) are sufficient. \smallskip

\noindent\textbf{Cross-LLM backbone robustness.} The cues target LLM-generation properties rather than a specific model or agent runtime. To test this, we run two additional persona-simulator backbones on OSMI alongside the Sonnet results above: \texttt{gpt-oss-120b} and \texttt{Gemini-3-Flash} (both via raw chat-completions, in contrast to Sonnet's Claude Code agent runtime). Despite this implementation asymmetry, SPABD $K{=}5$ holds across all three backbones (mean AUROC 0.90, within $\pm$0.02). Existing detectors are far less robust. SPABD beats the best per-cell baseline by $+0.25$ to $+0.34$ on every backbone (Tab.~\ref{tab:crossllm}, App.~\ref{sec:ablation_tables}). \smallskip

\noindent\textbf{Aggregator choice.} Since SPABD aggregates the four cue $z$-scores via a one-tail max for scoring, we study the effect of using different unsupervised aggregation strategies. We find that mean-$z$ and sum-$z$ perform slightly better on average under clean settings ($0.776$ vs.\ $0.751$), but we adopt max-$z$ for its robustness. It relies on whichever cue is most informative for a given respondent, allowing it to remain effective when other cues are weakened. For example, under a blind attack that varies answer lengths to appear more human-like, max-$z$ achieves $0.839$ while mean-$z$ drops to $0.722$. Overall, performance is relatively insensitive to the choice of aggregation strategy. The full results are reported in App.~\ref{sec:ablation_tables}. \smallskip

\noindent\textbf{The number of open-ended fields.} SPABD is designed for online surveys with multiple open-ended text fields. As the number of such fields decreases, detection performance declines gracefully. When half of the fields are removed, the average AUROC is reduced to 0.655, while still remaining superior to the performance of the best existing detector ($0.61$) when given all survey texts from the \emph{full} surveys (Tab.~\ref{tab:field_count}). This is expected because behavioural cues are most informative when observed across the entire survey and through the relationships between questions and their answers, which is exactly why SPABD performs better than applying MGT models to concatenated text. When the number of text fields is reduced to one, it becomes MGT detection over short text.

\noindent\textbf{Reference sample quality.} We analyse how the quality of the $K$ reference responses affects SPABD at inference time, considering both contamination and references drawn from other surveys. SPABD is relatively robust to mild contamination. Replacing one of five verified human responses with an AI response reduces mean AUROC from 0.751 to 0.730, while replacing two reduces it to 0.692, both still well above the best-performing baseline at 0.61. However, using reference responses from outside the target survey leads to a non-trivial reduction in performance. This setting is not intended by design. The verified responses should therefore be selected with caution. Full results are reported in App.~\ref{sec:ablation_tables}.

\section{Conclusion}
We introduce ASURRE, a dataset for AI-assisted online survey participation, on which existing MGT detection fails against persona-grounded agentic completion. Our analysis shows that although this sophisticated usage weakens textual signals, behavioural cues exploiting discrepancies between human variability and LLM regularities remain discriminative. To validate the promise of using such behavioural cues for detection, we introduce a simple yet effective baseline method, SPABD, that is training-free and uses few-shot aggregation of our identified behavioural cues. Across three surveys in different domains, it achieves a 0.14 AUROC improvement over the best existing baseline and remains effective under different usage strategies.

\clearpage

\section*{Limitations}

\textbf{Applicable settings.} SPABD requires a survey with multiple open-ended fields per respondent. This is the setting in which AI-assisted completion is both most likely and most consequential for validity. Open-ended fields are standard practice across most disciplines in the social sciences. At inference, a small set of verified human responses is needed to anchor the reference distribution. The cues are undefined for single-field instruments, and detection weakens as the number of open-ended fields decreases (App.~\ref{sec:ablation_tables}). Even though the default number of reference responses is small (5), removing them would limit the performance of SPABD.

\noindent \textbf{Resilience against adversarial attacks.} Our study concerns agentic survey completion with persona conditioning, where moderate engineering effort is spent on achieving human-like realism or mimicking a persona. However, malicious, targeted adversarial attacks are outside the scope of our evaluation and should be studied separately.

\noindent  \textbf{Effective but imperfect.} Although SPABD substantially outperforms existing zero-shot detectors on agentic survey completion, its performance remains imperfect under sophisticated agentic completion. Our work validates behavioural detection as a promising direction rather than a complete solution.

\noindent \textbf{Language coverage.} We evaluate persona-grounded LLM agents in four control modes on three primarily English-language surveys, among which PhD2019 includes some non-English responses. It is possible that behavioural cues may have different characteristics in other languages.

\section*{Ethical Considerations}

\textbf{Broader impact.} Since our work could make AI-assisted survey completion easier to carry out, we release only the high-level architecture of the agentic completion pipelines alongside our behavioural analyses, and will provide full implementation details on a case-by-case basis upon request. The detection framework is intended as a practical tool for researchers working with survey data. We are conscious that false detections, particularly false positives, carry costs for participants, and recommend that scores be used as screening signals rather than as grounds for exclusion or non-payment.

Participants or adversaries may attempt to evade SPABD's cues, since they are interpretable and fully described. Robustness of such adversarial attacks is not in the scope of our study, but it is important to note that all currently published detectors are not immune to this threat and, therefore, it is not unique to SPABD.

\noindent \textbf{Risk of demographic bias.} MGT models can systematically disadvantage non-native English writers \citep{liang2023gpt}. SPABD does not rely on fluency, perplexity, or vocabulary, and only one of its four cues is length-based. We nevertheless evaluate potential bias related to these factors to assess fairness. We compare cue values and flag rates between native and non-native respondents across all three surveys, using interface language or country of residence as proxies for respondent background (App. J). We find that non-native respondents do not show more uniform answer lengths on any survey. On the largest survey, their answers are also shorter and less closely aligned with the questions than those of native respondents. Both characteristics are less similar to the typical AI pattern than those observed in the native group. As a result, non-native respondents are flagged less often rather than more. Nevertheless, we recommend using flagged responses for human review rather than automatic rejection. Auditing flag rates across respondent groups is also recommended.

\section*{Acknowledgments}

Qizhou Wang, Bogdan Mamaev, and Christopher Leckie were supported in part by the Australian Internet Observatory (AIO), a national research infrastructure supporting digital platform and smart data research. AIO received investment from the Australian Research Data Commons (ARDC) through the National Collaborative Research Infrastructure Strategy (NCRIS).\\
This work was supported in part by the ARC Centre of Excellence on Automated Decision Making and Society CE200100005.\\
This research was supported by The University of Melbourne's Research Computing Services and the Petascale Campus Initiative.\\
This material is based upon work supported by the Google Cloud Research Credits program.

\bibliography{custom}

@misc{mitchell2023detectgpt,
      title={{DetectGPT}: Zero-Shot Machine-Generated Text Detection using Probability Curvature}, 
      author={Eric Mitchell and Yoonho Lee and Alexander Khazatsky and Christopher D. Manning and Chelsea Finn},
      year={2023},
      eprint={2301.11305},
      archivePrefix={arXiv},
      primaryClass={cs.CL},
      url={https://arxiv.org/abs/2301.11305}, 
}

@misc{hans2024binoculars,
      title={Spotting {LLM}s With {Binoculars}: Zero-Shot Detection of Machine-Generated Text}, 
      author={Abhimanyu Hans and Avi Schwarzschild and Valeriia Cherepanova and Hamid Kazemi and Aniruddha Saha and Micah Goldblum and Jonas Geiping and Tom Goldstein},
      year={2024},
      eprint={2401.12070},
      archivePrefix={arXiv},
      primaryClass={cs.CL},
      url={https://arxiv.org/abs/2401.12070}, 
}

@misc{bao2024fastdetectgpt,
      title={Fast-{DetectGPT}: Efficient Zero-Shot Detection of Machine-Generated Text via Conditional Probability Curvature}, 
      author={Guangsheng Bao and Yanbin Zhao and Zhiyang Teng and Linyi Yang and Yue Zhang},
      year={2024},
      eprint={2310.05130},
      archivePrefix={arXiv},
      primaryClass={cs.CL},
      url={https://arxiv.org/abs/2310.05130}, 
}

@misc{hu2023radar,
      title={RADAR: Robust {AI}-Text Detection via Adversarial Learning}, 
      author={Xiaomeng Hu and Pin-Yu Chen and Tsung-Yi Ho},
      year={2023},
      eprint={2307.03838},
      archivePrefix={arXiv},
      primaryClass={cs.CL},
      url={https://arxiv.org/abs/2307.03838}, 
}

@misc{veselovsky2023artificial,
      title={Artificial Artificial Artificial Intelligence: Crowd Workers Widely Use Large Language Models for Text Production Tasks}, 
      author={Veniamin Veselovsky and Manoel Horta Ribeiro and Robert West},
      year={2023},
      eprint={2306.07899},
      archivePrefix={arXiv},
      primaryClass={cs.CL},
      url={https://arxiv.org/abs/2306.07899}, 
}

@article{argyle2023one,
  title={Out of one, many: Using language models to simulate human samples},
  author={Argyle, Lisa P and Busby, Ethan C and Fulda, Nancy and Gubler, Joshua R and Rytting, Christopher and Wingate, David},
  journal={Political Analysis},
  volume={31},
  number={3},
  pages={337--351},
  year={2023},
  publisher={Cambridge University Press}
}

@misc{park2024generative,
      title={{LLM} Agents Grounded in Self-Reports Enable General-Purpose Simulation of Individuals}, 
      author={Joon Sung Park and Carolyn Q. Zou and Jonne Kamphorst and Niles Egan and Aaron Shaw and Benjamin Mako Hill and Carrie Cai and Meredith Ringel Morris and Percy Liang and Robb Willer and Michael S. Bernstein},
      year={2026},
      eprint={2411.10109},
      archivePrefix={arXiv},
      primaryClass={cs.AI},
      url={https://arxiv.org/abs/2411.10109}, 
}

@misc{reimers2019sbert,
      title={Sentence-{BERT}: Sentence Embeddings using {S}iamese {BERT}-Networks}, 
      author={Nils Reimers and Iryna Gurevych},
      year={2019},
      eprint={1908.10084},
      archivePrefix={arXiv},
      primaryClass={cs.CL},
      url={https://arxiv.org/abs/1908.10084}, 
}

@misc{wang2020minilm,
      title={{MiniLM}: Deep Self-Attention Distillation for Task-Agnostic Compression of Pre-Trained Transformers}, 
      author={Wenhui Wang and Furu Wei and Li Dong and Hangbo Bao and Nan Yang and Ming Zhou},
      year={2020},
      eprint={2002.10957},
      archivePrefix={arXiv},
      primaryClass={cs.CL},
      url={https://arxiv.org/abs/2002.10957}, 
}

@misc{solaiman2019release,
      title={Release Strategies and the Social Impacts of Language Models}, 
      author={Irene Solaiman and Miles Brundage and Jack Clark and Amanda Askell and Ariel Herbert-Voss and Jeff Wu and Alec Radford and Gretchen Krueger and Jong Wook Kim and Sarah Kreps and Miles McCain and Alex Newhouse and Jason Blazakis and Kris McGuffie and Jasmine Wang},
      year={2019},
      eprint={1908.09203},
      archivePrefix={arXiv},
      primaryClass={cs.CL},
      url={https://arxiv.org/abs/1908.09203}, 
}

@inproceedings{gehrmann2019gltr,
    title = "{GLTR}: Statistical Detection and Visualization of Generated Text",
    author = "Gehrmann, Sebastian  and
      Strobelt, Hendrik  and
      Rush, Alexander",
    editor = "Costa-juss{\`a}, Marta R.  and
      Alfonseca, Enrique",
    booktitle = "Proceedings of the 57th Annual Meeting of the Association for Computational Linguistics: System Demonstrations",
    month = jul,
    year = "2019",
    address = "Florence, Italy",
    publisher = "Association for Computational Linguistics",
    url = "https://aclanthology.org/P19-3019/",
    doi = "10.18653/v1/P19-3019",
    pages = "111--116"
}

@misc{wu2025survey,
      title={A Survey on {LLM}-Generated Text Detection: Necessity, Methods, and Future Directions}, 
      author={Junchao Wu and Shu Yang and Runzhe Zhan and Yulin Yuan and Derek F. Wong and Lidia S. Chao},
      year={2024},
      eprint={2310.14724},
      archivePrefix={arXiv},
      primaryClass={cs.CL},
      url={https://arxiv.org/abs/2310.14724}, 
}

@inproceedings{yang2023survey,
    title = "A Survey on Detection of {LLM}s-Generated Content",
    author = "Yang, Xianjun  and
      Pan, Liangming  and
      Zhao, Xuandong  and
      Chen, Haifeng  and
      Petzold, Linda Ruth  and
      Wang, William Yang  and
      Cheng, Wei",
    editor = "Al-Onaizan, Yaser  and
      Bansal, Mohit  and
      Chen, Yun-Nung",
    booktitle = "Findings of the Association for Computational Linguistics: EMNLP 2024",
    month = nov,
    year = "2024",
    address = "Miami, Florida, USA",
    publisher = "Association for Computational Linguistics",
    doi = "10.18653/v1/2024.findings-emnlp.572",
    pages = "9786--9805"
}

@misc{openai2025gptoss,
      title={gpt-oss-120b \& gpt-oss-20b Model Card}, 
      author={OpenAI},
      year={2025},
      eprint={2508.10925},
      archivePrefix={arXiv},
      primaryClass={cs.CL},
      url={https://arxiv.org/abs/2508.10925}, 
}

@misc{yang2025qwen3,
      title={Qwen3 Technical Report}, 
      author={An Yang and Anfeng Li and Baosong Yang and Beichen Zhang and Binyuan Hui and Bo Zheng and Bowen Yu and Chang Gao and Chengen Huang and Chenxu Lv and Chujie Zheng and Dayiheng Liu and Fan Zhou and Fei Huang and Feng Hu and Hao Ge and Haoran Wei and Huan Lin and Jialong Tang and Jian Yang and Jianhong Tu and Jianwei Zhang and Jianxin Yang and Jiaxi Yang and Jing Zhou and Jingren Zhou and Junyang Lin and Kai Dang and Keqin Bao and Kexin Yang and Le Yu and Lianghao Deng and Mei Li and Mingfeng Xue and Mingze Li and Pei Zhang and Peng Wang and Qin Zhu and Rui Men and Ruize Gao and Shixuan Liu and Shuang Luo and Tianhao Li and Tianyi Tang and Wenbiao Yin and Xingzhang Ren and Xinyu Wang and Xinyu Zhang and Xuancheng Ren and Yang Fan and Yang Su and Yichang Zhang and Yinger Zhang and Yu Wan and Yuqiong Liu and Zekun Wang and Zeyu Cui and Zhenru Zhang and Zhipeng Zhou and Zihan Qiu},
      year={2025},
      eprint={2505.09388},
      archivePrefix={arXiv},
      primaryClass={cs.CL},
      url={https://arxiv.org/abs/2505.09388}, 
}

@misc{google2025gemini3,
  title        = {{Gemini} 3 Flash},
  author       = {{Google DeepMind}},
  year         = {2025},
  howpublished = {\url{https://deepmind.google/technologies/gemini/}}
}

@misc{anthropic2025sonnet,
  title        = {{Claude Sonnet 4.6}},
  author       = {{Anthropic}},
  year         = {2025},
  howpublished = {\url{https://www.anthropic.com/claude}}
}

@article{meade2012identifying,
  title   = {Identifying careless responses in survey data},
  author  = {Meade, Adam W. and Craig, S. Bartholomew},
  journal = {Psychological Methods},
  volume  = {17},
  number  = {3},
  pages   = {437--455},
  year    = {2012}
}

@article{huang2012detecting,
  title   = {Detecting and deterring insufficient effort responding to surveys},
  author  = {Huang, Jason L. and Curran, Paul G. and Keeney, Jessica and Poposki, Elizabeth M. and DeShon, Richard P.},
  journal = {Journal of Business and Psychology},
  volume  = {27},
  number  = {1},
  pages   = {99--114},
  year    = {2012}
}

@article{kennedy2020shape,
  title   = {The Shape of and Solutions to the {MT}urk Quality Crisis},
  author  = {Kennedy, Ryan and Clifford, Scott and Burleigh, Tyler and Waggoner, Philip D. and Jewell, Ryan and Winter, Nicholas J. G.},
  journal = {Political Science Research and Methods},
  volume  = {8},
  number  = {4},
  pages   = {614--629},
  year    = {2020}
}

@misc{dugan2024raid,
      title={{RAID}: A Shared Benchmark for Robust Evaluation of Machine-Generated Text Detectors}, 
      author={Liam Dugan and Alyssa Hwang and Filip Trhlik and Josh Magnus Ludan and Andrew Zhu and Hainiu Xu and Daphne Ippolito and Chris Callison-Burch},
      year={2024},
      eprint={2405.07940},
      archivePrefix={arXiv},
      primaryClass={cs.CL},
      url={https://arxiv.org/abs/2405.07940}, 
}

@book{groves2011survey,
  title={Survey methodology},
  author={Groves, Robert M and Fowler Jr, Floyd J and Couper, Mick P and Lepkowski, James M and Singer, Eleanor and Tourangeau, Roger},
  year={2011},
  publisher={John Wiley \& Sons}
}

@article{krosnick1991response,
  title={Response strategies for coping with the cognitive demands of attitude measures in surveys},
  author={Krosnick, Jon A},
  journal={Applied cognitive psychology},
  volume={5},
  number={3},
  pages={213--236},
  year={1991},
  publisher={Wiley Online Library}
}

@article{zhang2025generative,
  title={Generative {AI} meets open-ended survey responses: Research participant use of {AI} and homogenization},
  author={Zhang, Simone and Xu, Janet and Alvero, AJ},
  journal={Sociological Methods \& Research},
  volume={54},
  number={3},
  pages={1197--1242},
  year={2025},
  publisher={SAGE Publications Sage CA: Los Angeles, CA}
}

@article{liang2023gpt,
  title={{GPT} detectors are biased against non-native English writers},
  author={Liang, Weixin and Yuksekgonul, Mert and Mao, Yining and Wu, Eric and Zou, James},
  journal={Patterns},
  volume={4},
  number={7},
  year={2023},
  publisher={Elsevier}
}

\appendix

\section{AI Response Simulation}
\label{sec:simulation}

\subsection{Source Datasets}
\label{sec:datasets}

We use three publicly available survey datasets, all collected before widespread LLM availability (pre-2020), so that every original response is verifiably human-written and free of AI contamination. Tab.~\ref{tab:datasets} summarises the corpora.

\begin{table}[h]
\centering
\begin{adjustbox}{max width=\columnwidth}
\begin{tabular}{lcrrrr}
\toprule
Dataset & Year & Raw & Retained & Train & Test \\
\midrule
phd2019        & 2019     & 6{,}813 & 6{,}813 & 5{,}450 & 1{,}363 \\
springer2017 & 2017     & 5{,}292 & 1{,}596 & 1{,}276 & \phantom{0,}320 \\
osmi\_mh       & 2017--18 & 1{,}173 & \phantom{0,}534 & \phantom{0,}427 & \phantom{0,}107 \\
\bottomrule
\end{tabular}
\end{adjustbox}
\caption{Source datasets. \emph{Raw}: original number of participants. \emph{Retained}: respondents kept after eligibility filtering ($\geq 80$ characters of open-text content).}
\label{tab:datasets}
\end{table}

\textbf{phd2019.} A global survey of PhD students on experiences, satisfaction, supervisor relationships, mental health, and career intentions, conducted by \textit{Nature} / Shift Learning (CC BY 4.0; Figshare DOI \texttt{10.6084/m9.figshare.10266299}). We use the six open-text fields listed in Tab.~\ref{tab:phd2019_fields}. Median 26 words per respondent (IQR 10--61); Q26 is the richest field and Q17 is uniformly very short ($\leq 15$ words).

\begin{table}[h]
\centering
\small
\begin{tabular}{@{}lp{0.78\columnwidth}@{}}
\toprule
ID & Question \\
\midrule
Q26 & How would you describe the academic system based on your PhD experience? \\
Q16 & Is there anything else that has concerned you since starting your PhD? \\
Q55 & What one thing do you wish you'd known when you started your PhD? \\
Q41 & What type of career are you interested in pursuing after your degree? \\
Q60 & Any further comments? \\
Q17 & What do you enjoy most about life as a PhD student? \\
\bottomrule
\end{tabular}
\caption{Open-text fields used from phd2019.}
\label{tab:phd2019_fields}
\end{table}

\textbf{springer2017.} A global survey of academic researchers on professional use of social media and scholarly collaboration networks, run by Springer Nature in 2017 (CC BY 4.0; Figshare DOI \texttt{10.6084/m9.figshare.5028212}). Of 5{,}292 raw respondents, 1{,}596 contributed sufficient open-text content. We use the seven open-text fields listed in Tab.~\ref{tab:springer2017_fields}. Responses are short and telegraphic (median 9 words per respondent, IQR 4--27; most fields have a median of 1--3 words).

\begin{table}[h]
\centering
\small
\begin{tabular}{@{}lp{0.78\columnwidth}@{}}
\toprule
ID & Question \\
\midrule
Q23 & Other tasks you use social media for in relation to your work. \\
Q29 & Comments on usefulness of social media for supporting research. \\
Q37 & Comments on usefulness of social media for sharing professional content. \\
Q43 & Comments on usefulness of social media for promoting professional content. \\
Q47 & Comments on usefulness of social media for networking and collaboration. \\
Q63 & What could publishers provide or improve to help your use of social media? \\
Q65 & Further comments on professional use of social media. \\
\bottomrule
\end{tabular}
\caption{Open-text fields used from springer2017.}
\label{tab:springer2017_fields}
\end{table}

\textbf{osmi\_mh.} The annual Mental Health in Tech Survey from Open Sourcing Mental Illness, combining the 2017 and 2018 releases (CC BY-SA 4.0; \texttt{osmihelp.org}). Of 1{,}173 raw respondents, 534 are retained. We use ten open-text fields (Tab.~\ref{tab:osmi_fields}); the five narrative fields covering current/previous employer and coworker conversations (Q\_EC, Q\_CC, Q\_CW, Q\_PE, Q\_PC) carry the bulk of the content. Responses are the longest and most narrative-rich in our study (median 45 words per respondent, IQR 26--79), and the only ones centred on a sensitive personal topic.

\begin{table}[h]
\centering
\small
\begin{tabular}{@{}lp{0.78\columnwidth}@{}}
\toprule
ID & Question \\
\midrule
Q\_EC & Conversation with current employer about mental health. \\
Q\_CC & Conversation with current coworkers about mental health. \\
Q\_CW & Coworker's disclosure of their mental health to you. \\
Q\_PE & Conversation with previous employer about mental health. \\
Q\_PC & Conversation with previous coworkers about mental health. \\
Q\_CA & How being identified as having a mental health issue affected your career. \\
Q\_BH & Circumstances of a badly handled or unsupportive workplace response. \\
Q\_SH & Circumstances of a supportive or well-handled workplace response. \\
Q\_ID & What could the industry or employers do to improve mental health support? \\
Q\_AE & Any additional comments. \\
\bottomrule
\end{tabular}
\caption{Open-text fields used from osmi\_mh.}
\label{tab:osmi_fields}
\end{table}

\subsection{Simulation Patterns}
\label{sec:patterns}

We simulate three patterns of LLM use. All three operate at the respondent level: the unit is a complete answer set, not an isolated field.

\subsubsection{LLM Revision}
The respondent writes their own answers and runs each through an LLM purely for stylistic polishing. The LLM receives the raw answer text, with no question label or surrounding fields, and is instructed to rewrite the prose while preserving meaning, opinions, and concrete details. Fields are processed independently as an asynchronous batch via vLLM.

\subsubsection{Full Generation}
The respondent delegates authoring entirely to an LLM. The LLM receives only the question label and is prompted to respond as a researcher quickly completing the survey. Because no human input or respondent context is provided, outputs are plausible but ungrounded. Fields are again processed independently.

\subsubsection{Persona Simulation}
A two-stage pipeline mimics a specific real respondent. \textit{Stage A (portrait)}: Claude reads the respondent's full answer set and produces a structured persona portrait (Section~\ref{sec:portrait}). \textit{Stage B (simulation)}: a fresh Claude agent session fills the survey one question per turn, conditioned on the portrait and accumulating prior answers in context. Three skills regulate generation: \texttt{ground\_persona} maps each portrait dimension to concrete writing behaviour before the first answer; \texttt{survey\_fill} produces each answer under length and consistency constraints; and \texttt{humanise} performs a post-generation pass that suppresses AI stylistic markers and injects register-appropriate informality. Unlike the field-independent patterns, this produces internally consistent, persona-grounded answer sets.

The agent autonomously decides which questions to answer; skipping is signalled by emitting exactly \texttt{[SKIP]}, which is excluded from the output. We evaluate two engagement modes. The \emph{natural} variant (\texttt{persona\_sonnet}) lets the agent choose purely from its portrait, with no external policy. The \emph{controlled} variant (\texttt{persona\_sonnet\_controlled}) additionally applies a deterministic policy that pre-marks some questions as must-skip based on the persona's seriousness and a per-question tier (mandatory / standard / optional), with an eligibility floor of at least one answered question per persona. The persona portrait is the only carry-over from the source respondent into the simulation --- no field-level content or coverage is propagated.

\subsubsection{Persona Portrait Dimensions}
\label{sec:portrait}

The portrait specifies eight dimensions inferred from the source respondent's full answer set (Tab.~\ref{tab:portrait}). All categorical dimensions are inferred by an LLM; \texttt{response\_length\_habit} is computed directly from the respondent's empirical answer lengths, providing a quantitative anchor on response scale that the LLM does not invent.

\begin{table*}[h]
\centering
\small
\begin{adjustbox}{max width=\textwidth}
\begin{tabular}{lll}
\toprule
Dimension & Values & Role \\
\midrule
\texttt{affect}                & positive / negative / mixed / neutral        & emotional register of every answer \\
\texttt{engagement\_level}     & high / moderate / low                        & depth of engagement (survey-dependent) \\
\texttt{english\_level}        & native / fluent / proficient / intermediate  & vocabulary range, idiom use \\
\texttt{grammar\_error\_level} & none / minor / moderate / frequent           & error rate, applied consistently \\
\texttt{survey\_seriousness}   & high / moderate / low                        & effort; drives controlled-mode rates \\
\texttt{writing\_style}        & free text (2 sentences)                      & sentence length, formality, patterns \\
\texttt{main\_themes}          & 2--3 topics                                   & topics that recur across answers \\
\texttt{response\_length\_habit} & derived statistic                          & data-grounded length anchor \\
\bottomrule
\end{tabular}
\end{adjustbox}
\caption{Persona portrait dimensions used to control agentic completion.}
\label{tab:portrait}
\end{table*}

\paragraph{Per-dataset portrait distributions.} Inferred portraits over the eligible respondents in each dataset reveal markedly different population profiles:
\begin{itemize}\setlength\itemsep{0pt}
\item \textbf{phd2019} ($n{=}100$): \emph{affect} mixed 79\%, negative 15\%, positive 5\%; \emph{strain\_level} medium 57\%, high 34\%, low 9\%; \emph{survey\_seriousness} moderate 57\%, low 23\%, high 20\%; \emph{english\_level} proficient 45\%, native 26\%, fluent 21\%, intermediate 8\%.
\item \textbf{springer2017} ($n{=}90$): \emph{affect} positive 52\%, neutral 23\%, mixed 17\%, negative 8\%; \emph{engagement\_level} low 62\%, moderate 37\%, high 1\%; \emph{survey\_seriousness} low 61\%, moderate 32\%, high 7\%; \emph{english\_level} proficient 66\%, fluent 16\%, native 11\%, intermediate 8\%.
\item \textbf{osmi\_mh} ($n{=}107$): \emph{affect} predominantly mixed/negative (disclosure-anxiety register); \emph{disclosure\_comfort} spread across low/medium/high; \emph{survey\_seriousness} moderate-to-high (self-selected respondents); \emph{english\_level} mostly native/fluent.
\end{itemize}

\subsection{Models}
LLM Revision and Full Generation use the open-weight models \texttt{gpt-oss-120b} and \texttt{Qwen3.5-35B-A3B} and the proprietary Gemini 3 Flash Preview. Persona Simulation uses Claude Sonnet 4.6 with both the natural and controlled engagement modes described above. This combination spans open-weight and commercial families across multiple providers, allowing us to assess detection generalisation across model sources. All splits use balanced human and AI samples following the partitions in Tab.~\ref{tab:datasets}.

\section{Implementation Details}
\subsection{Detection Methods}
\label{sec:detectors}
We evaluate six AI-text detectors covering the main families of zero-shot detection plus one supervised baseline. All are run zero-shot on our survey corpora --- no fine-tuning on survey text. We summarise each below.

\paragraph{Zero-Shot Prompt Classifier (\texttt{zeroshot}).} Direct prompt-based binary classification: \texttt{gpt-oss-120b} is asked whether each input is AI-generated, and the binary label is parsed from a JSON reply (\texttt{temperature=0}).

\paragraph{Fast-DetectGPT (\texttt{fdgpt}).} Sampling-free curvature score using a single model's logits to construct a per-token reference distribution \citep{bao2024fastdetectgpt}. Scored with \texttt{Qwen3-4B}.

\paragraph{DetectGPT (\texttt{dgpt}).} The original perturbation-based curvature detector \citep{mitchell2023detectgpt}: ten mask-fill perturbations per input; AI text appears as a local maximum in the log-probability landscape. Scored with \texttt{Qwen3-4B}.

\paragraph{Log-Likelihood (\texttt{loglik}).} The simplest baseline: per-token mean log-probability under \texttt{Qwen3-4B}. AI-sampled text tends to sit in higher-probability regions than human text.

\paragraph{Binoculars (\texttt{bino}).} Cross-perplexity ratio between a base (observer) and instruction-tuned (performer) model \citep{hans2024binoculars}. We use \texttt{Qwen3-4B} as observer and \texttt{Qwen3-4B-Instruct-2507} as performer.

\paragraph{RADAR (\texttt{radar}).} The only supervised detector in the matrix: an adversarially-trained RoBERTa-style classifier (\texttt{TrustSafeAI/RADAR-Vicuna-7B}) producing a softmax probability for the AI class \citep{hu2023radar}.

\subsection{Behavioural Cue Computation}
\label{sec:cue_computation}
Let respondent $r$ have answered set $A_r = \{(q_i, a_i)\}_{i=1}^{K_r}$, where $q_i$ is the question prompt and $a_i$ the free-text answer. Let $\ell_i = |a_i|$ be the whitespace-split word count, and $\phi(\cdot)$ the L2-normalised sentence encoder (\texttt{sentence-transformers/all-MiniLM-L6-v2}). The four cues are
\begin{align*}
\texttt{min\_len}(r)      &= \min_{i} \ell_i, \\
\texttt{length\_cv}(r)    &= \mathrm{std}_i(\ell_i) \,/\, \mathrm{mean}_i(\ell_i), \\
\texttt{mean\_qa\_sim}(r) &= \tfrac{1}{K_r}\sum_{i=1}^{K_r} \phi(q_i) \cdot \phi(a_i), \\
\texttt{n\_answered}(r)   &= K_r.
\end{align*}
All cues depend only on $r$'s own answers and the survey question texts --- no reference language model, no labels, no peer respondents. The encoder is 80M parameters and runs once per respondent.

\subsection{Evaluation Metrics}
We report AUROC, the area under the receiver operating characteristic curve. Values range in $[0,1]$: 1 indicates perfect ranking (every AI response scored above every human response), 0.5 indicates random ranking. AUROC is invariant to monotonic transformations of $s(\cdot)$ and independent of any decision threshold, making it a measure of discriminative ability rather than operating-point performance.

\section{Question--Answer Homogenisation}
\label{sec:ols_homog}

\begin{figure}[t]
\centering
\includegraphics[width=\columnwidth]{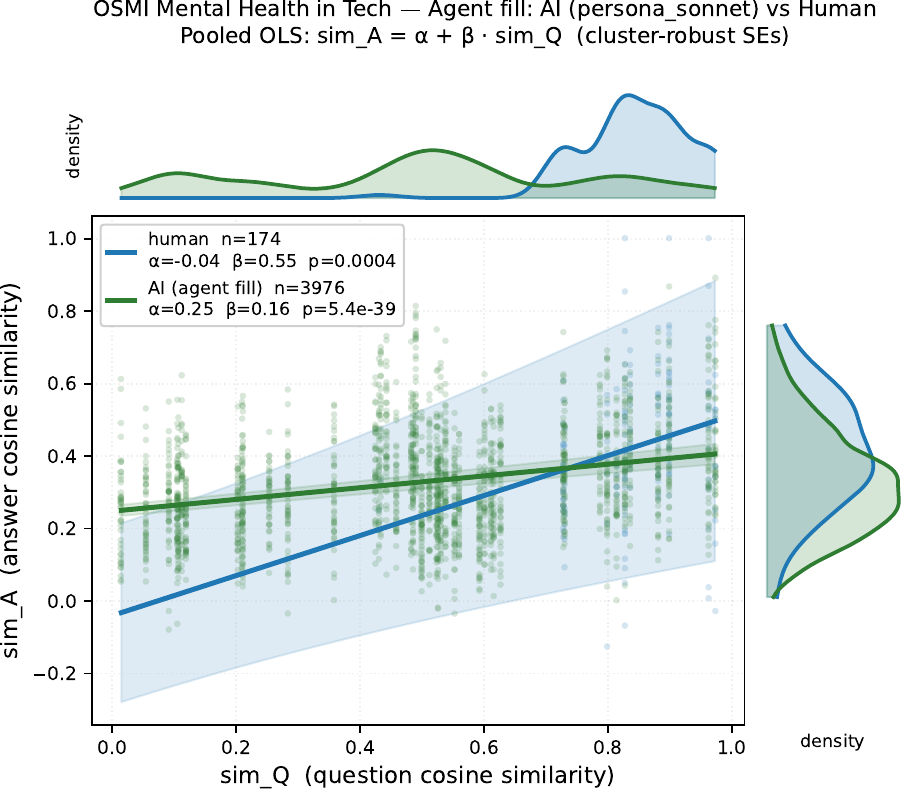}
\caption{Within-respondent question--answer similarity on OSMI \texttt{agent\_fill}. Humans (left) track question similarity ($\beta=0.55$); persona AI (right) collapses to a portrait-anchored register ($\beta=0.16$; non-overlapping CIs).}
\label{fig:ols_homog}
\end{figure}

We probe the question--answer coupling behind \texttt{mean\_qa\_sim} on OSMI agent\_fill. For each respondent we form within-respondent pairs of (question similarity, answer similarity) across their answered fields and fit $\mathrm{sim}_A = \alpha + \beta\,\mathrm{sim}_Q$ separately for the human and persona-AI pools (Fig.~\ref{fig:ols_homog}). Human respondents track question similarity ($\beta = 0.55$): when two questions are semantically close, their answers are too. Persona-grounded AI completions homogenise into a narrower, less question-coupled distribution ($\beta = 0.16$, non-overlapping 95\% CIs), reflecting the portrait-anchored compression of $\mathrm{sim}_A$ visible in the marginal densities.

\section{Detailed Experiment Results}
\label{sec:full_results}

Tab.~\ref{tab:auroc_full} reports the full per-augmenter test-split AUROC matrix expanding the per-pattern averages in Tab.~\ref{tab:main_results} (main paper).

\begin{table*}[t]
\centering
\begin{adjustbox}{max width=\textwidth}
\begin{tabular}{ll *{3}{c} *{3}{c} *{4}{c}}
\toprule
& & \multicolumn{3}{c}{LLM Revision} & \multicolumn{3}{c}{Full Generation} & \multicolumn{4}{c}{Agentic Fill} \\
\cmidrule(lr){3-5} \cmidrule(lr){6-8} \cmidrule(lr){9-12}
Dataset & Method & OSS120 & Qwen3.5 & Gemini3 & OSS120 & Qwen3.5 & Gemini3 & \textsc{nat} & \textsc{ctrl} & \textsc{len} & \textsc{cmb} \\
\midrule
\multirow{6}{*}{phd2019}
 & Zero-shot   & 0.623 & 0.595 & 0.664 & 0.465 & 0.424 & 0.445 & 0.428 & 0.447 & 0.442 & 0.463 \\
 & Fast-DGPT   & 0.613 & 0.616 & 0.693 & 0.833 & 0.920 & 0.845 & 0.498 & 0.550 & 0.538 & 0.527 \\
 & DetectGPT   & 0.549 & 0.555 & 0.590 & 0.809 & 0.911 & 0.779 & 0.609 & 0.593 & 0.562 & 0.569 \\
 & Log-Lik     & 0.654 & 0.616 & 0.718 & 0.885 & 0.949 & 0.878 & 0.562 & 0.573 & 0.572 & 0.578 \\
 & Binoculars  & 0.571 & 0.513 & 0.622 & 0.657 & 0.708 & 0.694 & 0.625 & 0.593 & 0.625 & 0.608 \\
 & RADAR       & 0.285 & 0.354 & 0.226 & 0.241 & 0.374 & 0.144 & 0.283 & 0.344 & 0.237 & 0.228 \\
\midrule
\multirow{6}{*}{springer2017}
 & Zero-shot   & 0.563 & 0.563 & 0.681 & 0.460 & 0.440 & 0.443 & 0.473 & 0.405 & 0.508 & 0.520 \\
 & Fast-DGPT   & 0.717 & 0.677 & 0.818 & 0.846 & 0.912 & 0.884 & 0.575 & 0.622 & 0.517 & 0.526 \\
 & DetectGPT   & 0.763 & 0.527 & 0.642 & 0.834 & 0.941 & 0.820 & 0.605 & 0.639 & 0.505 & 0.556 \\
 & Log-Lik     & 0.544 & 0.565 & 0.815 & 0.919 & 0.961 & 0.923 & 0.619 & 0.683 & 0.519 & 0.557 \\
 & Binoculars  & 0.371 & 0.405 & 0.553 & 0.667 & 0.677 & 0.709 & 0.557 & 0.554 & 0.562 & 0.586 \\
 & RADAR       & 0.375 & 0.416 & 0.224 & 0.345 & 0.594 & 0.221 & 0.424 & 0.510 & 0.345 & 0.312 \\
\midrule
\multirow{6}{*}{osmi\_mh}
 & Zero-shot   & 0.568 & 0.593 & 0.729 & 0.533 & 0.461 & 0.463 & 0.448 & 0.439 & 0.452 & 0.451 \\
 & Fast-DGPT   & 0.601 & 0.686 & 0.628 & 0.857 & 0.813 & 0.695 & 0.456 & 0.476 & 0.439 & 0.461 \\
 & DetectGPT   & 0.490 & 0.598 & 0.385 & 0.896 & 0.941 & 0.604 & 0.599 & 0.648 & 0.554 & 0.563 \\
 & Log-Lik     & 0.634 & 0.686 & 0.550 & 0.867 & 0.849 & 0.627 & 0.447 & 0.480 & 0.399 & 0.439 \\
 & Binoculars  & 0.545 & 0.593 & 0.508 & 0.747 & 0.592 & 0.705 & 0.533 & 0.510 & 0.554 & 0.522 \\
 & RADAR       & 0.431 & 0.417 & 0.349 & 0.646 & 0.791 & 0.429 & 0.560 & 0.588 & 0.497 & 0.525 \\
\bottomrule
\end{tabular}
\end{adjustbox}
\caption{Full test-split \textbf{AUROC} per (dataset, usage pattern, augmenter, detector). \textsc{nat}/\textsc{ctrl}/\textsc{len}/\textsc{cmb} = natural / controlled / length-restrictive / combined.}
\label{tab:auroc_full}
\end{table*}

\section{Ablation Studies}
\label{sec:ablation_tables}

This appendix backs the two ablations in Sec.~\ref{sec:ablations}. Tab.~\ref{tab:k_scan} reports SPABD as a function of the known-human reference size $K \in \{5, 10, 20, 50, 100\}$, averaged over the 12 agentic-completion cells and 50 random K-shot draws per cell. Mean AUROC climbs from 0.75 at $K{=}5$ to 0.79 at $K{=}100$, and the seed-level standard deviation falls from 0.09 to 0.02 -- $K{=}5$ captures most of the achievable AUROC, and the variance reduction at larger $K$ comes from a more stable reference distribution rather than from new signal. Tab.~\ref{tab:crossllm} reports the cross-LLM backbone ablation on OSMI agent\_fill (natural mode). The Sonnet column is the same persona-simulator run used in the main results above (Claude Code agent runtime); we add two more backbones, \texttt{gpt-oss-120b} and \texttt{Gemini-3-Flash}, run via raw chat-completions. SPABD $K{=}5$ holds at $0.886$/$0.909$/$0.895$ (within $\pm 0.02$), while the best per-cell baseline changes identity between backbones (DetectGPT $\to$ RADAR $\to$ Binoculars) and Fast-DetectGPT / Log-Likelihood fall below chance on \texttt{gpt-oss-120b}.

\begin{table}[h]
\centering
\small
\adjustbox{max width=\columnwidth}{%
\begin{tabular}{lcccccc}
\toprule
$K$           & 3 & 5    & 10   & 20   & 50   & 100  \\
\midrule
Mean AUROC    & 0.74 & 0.75 & 0.78 & 0.78 & 0.79 & 0.79 \\
Seed std      & 0.10 & 0.09 & 0.06 & 0.05 & 0.03 & 0.02 \\
\bottomrule
\end{tabular}%
}
\caption{SPABD scaling with $K$ (known-human reference size). Mean over 12 cells $\times$ 50 seeds.}
\label{tab:k_scan}
\end{table}

\begin{table}[t]
\centering\small
\caption{Cross-LLM ablation on OSMI agent\_fill (natural mode). AUROC of six existing detectors and our SPABD ($K{=}5$) across three persona-simulator backbones. Best per-cell baseline changes identity between backbones and Fast-DGPT and Log-Lik fall below chance on gpt-oss. SPABD generalises within $\pm 0.02$ and beats the best per-cell baseline by $+0.25$ to $+0.34$.}
\label{tab:crossllm}
\begin{adjustbox}{max width=\columnwidth}
\begin{tabular}{l ccc}
\toprule
Method & Sonnet 4.6 & gpt-oss-120b & Gemini-3-Flash \\
\midrule
Zero-shot & 0.448 & 0.456 & 0.481 \\
Fast-DGPT & 0.456 & 0.264 & 0.372 \\
DetectGPT & 0.599 & 0.517 & 0.497 \\
Log-Lik & 0.447 & 0.289 & 0.365 \\
Binoculars & 0.533 & 0.535 & 0.553 \\
RADAR & 0.560 & 0.657 & 0.455 \\
\midrule
Mean (6 detectors) & 0.507 & 0.453 & 0.454 \\
Best per cell & \textbf{0.599} & \textbf{0.657} & \textbf{0.553} \\
\midrule
\textbf{SPABD K=5 (ours)} & \textbf{0.886} & \textbf{0.909} & \textbf{0.895} \\
$\Delta$ vs best baseline & $+0.287$ & $+0.252$ & $+0.342$ \\
\bottomrule
\end{tabular}
\end{adjustbox}
\end{table}

\noindent\textbf{Aggregator choice.} Tab.~\ref{tab:aggregator-ablation} compares SPABD's max-$z$ against seven alternative unsupervised aggregators on the 12 agent\_fill cells. Z-score-based methods (max-$z$, sum-$z$, mean-$z$, trimmed-sum-$z$) use the same $K{=}5$ robust reference as SPABD; covariance- and density-based methods (robust Mahalanobis, Isolation Forest, One-Class SVM, LOF) use $K{=}20$, since $K{=}5$ is too small for stable 4-dimensional covariance or density estimation. We also include a fully supervised K=5+K=5 logistic classifier as an upper bound (mean AUROC 0.856, requires AI labels). Among zero-shot aggregators, max-$z$, sum-$z$, mean-$z$, robust Mahalanobis, and Isolation Forest fall within 0.03 of each other (mean AUROCs 0.751--0.776); trimmed-sum-$z$, One-Class SVM and LOF trail at 0.682--0.689 in the small-$K$, 4-dimensional regime. We adopt max-$z$ for two reasons: (i) each flag is traceable to a specific cue, supporting human review of decisions; (ii) it is robust to per-cell direction inversion of any single cue (e.g., \texttt{mean\_qa\_sim} on OSMI), where sum-$z$ or mean-$z$ would carry the inverted z into the score. The zero-shot max-$z$ method closes 57\% of the gap between existing zero-shot baselines (mean 0.61) and the supervised upper bound (0.86), without using any AI labels.

The clean average is not the whole picture. On osmi\_mh with the Gemini-3-Flash backbone we also score a \emph{blind} attacker, prompted to leave one to three questions blank and to vary answer lengths so they look natural, with no persona document and no access to any real respondent statistics (App.~\ref{sec:patterns}). Mean-$z$ is marginally ahead of max-$z$ without the attack ($0.899$ vs $0.895$) but falls to $0.722$ under it, while max-$z$ retains $0.839$. The reason is visible in the individual cues: the attacker's guessed length distribution inverts both length cues (\texttt{min\_len} $0.701 \to 0.459$, \texttt{length\_cv} $0.702 \to 0.383$), but its guessed skip rate still under-shoots real non-participation, so \texttt{n\_answered} remains at $1.000$. A one-tail max rides that single surviving cue; an average dilutes it across four. Because the blind attacker is generated without persona grounding, this pair is not a clean ablation of the attack alone, but the two aggregators are compared on identical generations, which is what the design choice turns on.

\begin{table}[t]
\centering\small
\caption{Aggregator ablation: SPABD's max-$z$ vs alternative unsupervised aggregators on the 12 agent\_fill cells. Z-based methods use a $K{=}5$ robust reference with the same reference draws as Tab.~\ref{tab:spabd_perf}; covariance- and density-based methods use $K{=}20$ ($K{=}5$ is too small for covariance estimation). Sum-$z$ is a monotone transform of mean-$z$ over the same four cues, so their AUROCs coincide by construction. 50 seeds, AUROC mean across 12 cells.}
\label{tab:aggregator-ablation}
\begin{adjustbox}{max width=\columnwidth}
\begin{tabular}{l ccc}
\toprule
Method & Mean & Min & Max \\
\midrule
Max-z (K=5, ours) & \textbf{0.751} & 0.557 & 0.886 \\
Sum-z (K=5) & {0.776} & 0.636 & 0.862 \\
Mean-z (K=5) & {0.776} & 0.636 & 0.862 \\
Trimmed-sum-z (K=5, drop top $|z|$) & {0.689} & 0.404 & 0.856 \\
Robust Mahalanobis (K=20) & {0.759} & 0.466 & 0.997 \\
Isolation Forest (K=20) & {0.756} & 0.566 & 0.906 \\
One-Class SVM (K=20) & {0.682} & 0.396 & 0.949 \\
Local Outlier Factor (K=20) & {0.682} & 0.504 & 0.831 \\
Logistic (5 humans + 5 AI; SUPERVISED) & {0.856} & 0.651 & 0.989 \\
\bottomrule
\end{tabular}
\end{adjustbox}
\end{table}

\noindent\textbf{Open-field count.} SPABD reads behaviour across a respondent's answered
fields, so its signal depends on how many open-ended fields a survey has. Tab.~\ref{tab:field_count}
retains the first 25/50/75/100\% of each survey's fields and recomputes the cues. Detection degrades
gracefully rather than collapsing: at half the fields SPABD averages $0.655$, still above the $0.61$
achieved by the best existing detector on \emph{full} surveys, and even at a quarter of the fields it
averages $0.612$. The trend is survey-dependent --- phd2019 and osmi\_mh lose most, while
springer2017 is nearly flat --- and because fields are retained as a prefix rather than sampled, part
of that variation reflects which fields survive rather than how many.

\noindent\textbf{Reference-set robustness.} The $K$ known-human responses are SPABD's only
supervision, so we test what happens when that anchor is imperfect (Tab.~\ref{tab:reference_robustness}).
Contamination is mild: replacing one of the five references with an AI response costs $0.021$ AUROC
($0.751 \to 0.730$) and replacing two costs $0.059$, both still well above the best existing detector.
Trimming the anchor recovers roughly half of that loss ($0.740$ and $0.716$), so a practitioner who
suspects their reference batch is impure should trim rather than enlarge it.

Transfer across surveys, however, does \emph{not} hold. Drawing the reference from a different
survey lowers the mean from $0.751$ to $0.646$ and roughly doubles the spread across cells (sd $0.088
\to 0.135$, range $0.393$--$0.910$); a cross-survey reference exceeds the $0.61$ baseline in only 14 of
24 cells. The cost is uneven --- $-0.030$ on phd2019 but $-0.227$ on osmi\_mh, whose respondents answer
far fewer fields than either other population, so an out-of-survey anchor misplaces the
\texttt{n\_answered} scale that carries most of its signal. We therefore report this as a requirement
rather than a robustness result: the reference set must come from the survey being screened. Since
$K{=}5$ suffices (Tab.~\ref{tab:k_scan}) and tolerates contamination, this is a light requirement in
practice --- a small verified pilot batch, which is standard survey design --- but it is a real one.

\noindent\textbf{Flagged design choices.} Review raised three settings that were fixed
rather than tuned. Tab.~\ref{tab:design_choices} varies each in turn. When a cue's MAD is
near zero on the reference sample we floor it; the alternatives are a fixed $\epsilon$
($0.768$), an IQR-based scale ($0.742$), or discarding the degenerate cue altogether, which
costs the most ($0.693$) because the discarded cue is often the one carrying the signal for
that survey. The minimum-character floor used when selecting a respondent's own text is
likewise not critical: over $40$--$160$ characters the mean moves within $0.746$--$0.765$
without an interior optimum, so the setting trades pool size against per-respondent evidence
rather than tuning the score. Finally, the DetectGPT baseline is insensitive to its
perturbation count: across $10$, $25$, $50$ and $100$ perturbations on a matched osmi\_mh
reproduction its AUROC varies by less than $0.002$, so the number of perturbations does not
explain that detector's standing in Tab.~\ref{tab:main_results}.

\begin{table}[t]
\centering\tiny
\caption{Sensitivity to the design choices raised in review, mean AUROC over the 12 agentic cells. \emph{Top}: handling of a near-zero MAD for a cue. \emph{Bottom}: minimum-character floor applied when selecting a respondent's own text.}
\label{tab:design_choices}
\begin{tabular}{lcccc}
\toprule
MAD$\,{\approx}\,0$ handling & floor (ours) & $\epsilon$ & IQR & discard cue \\
\midrule
Mean AUROC & \textbf{0.751} & 0.768 & 0.742 & 0.693 \\
\bottomrule
\end{tabular}

\vspace{0.6em}

\begin{tabular}{lcccccc}
\toprule
Character floor & 40 & 60 & 80 & 100 & 120 & 160 \\
\midrule
Mean AUROC & 0.746 & 0.751 & 0.759 & 0.760 & 0.758 & 0.765 \\
\bottomrule
\end{tabular}
\end{table}

\begin{table}[t]
\centering\small
\caption{SPABD ($K{=}5$) as open-ended fields are removed, mean AUROC over the four agentic modes. Fields are retained as a prefix of the survey's field order; absolute counts are 2/3/5/6 (phd2019), 2/4/6/7 (springer2017) and 3/5/8/10 (osmi\_mh).}
\label{tab:field_count}
\begin{tabular}{lcccc}
\toprule
Fields retained & 25\% & 50\% & 75\% & 100\% \\
\midrule
phd2019      & 0.544 & 0.604 & 0.622 & 0.753 \\
springer2017 & 0.680 & 0.690 & 0.670 & 0.687 \\
osmi\_mh     & 0.611 & 0.671 & 0.775 & 0.813 \\
\midrule
Mean         & 0.612 & 0.655 & 0.689 & \textbf{0.751} \\
\bottomrule
\end{tabular}
\end{table}

\begin{table}[t]
\centering\small
\caption{Reference-set robustness, mean AUROC over the 12 agentic cells. \emph{Top}: known-human reference responses replaced by AI ones. \emph{Bottom}: reference drawn from a different survey (rows: test survey; columns: reference survey).}
\label{tab:reference_robustness}
\begin{tabular}{lccc}
\toprule
Contaminated (of $K{=}5$) & 0 & 1 & 2 \\
\midrule
Plain anchor           & 0.751 & 0.730 & 0.692 \\
Trimmed anchor         & ---   & 0.740 & 0.716 \\
\bottomrule
\end{tabular}

\vspace{0.6em}

\begin{tabular}{lccc}
\toprule
Test $\backslash$ Ref. & phd2019 & springer2017 & osmi\_mh \\
\midrule
phd2019      & \textbf{0.753} & 0.712 & 0.733 \\
springer2017 & 0.662 & \textbf{0.687} & 0.599 \\
osmi\_mh     & 0.592 & 0.579 & \textbf{0.813} \\
\bottomrule
\end{tabular}
\end{table}

\section{Held-Out Validation}
\label{sec:loso}

The four cues and their sign priors were fixed from general properties of LLM generation
--- verbosity, length uniformity, over-coverage and on-topic-ness --- and applied unchanged to
every survey, model and prompting mode; Fig.~\ref{fig:orthogonality} is an oracle analysis of
their individual discriminability, not a selection step. To test whether that claim survives a
stricter protocol, we run a leave-one-survey-out study. For each held-out survey we fit a
configuration on the other two --- cues, sign directions and aggregator chosen by greedy
forward search over an $18$-feature pool --- freeze it, and evaluate it on the held-out survey,
whose data the search never sees. This is deliberately not our method; it is the fitted
alternative.

Tab.~\ref{tab:loso} reports the result. The fitted configuration beats the best existing
detector on every fold (mean $0.668$ vs $0.61$), which is itself evidence that respondent-level
behaviour carries signal. But it scores \emph{below} our fixed priors on every fold
($0.668$ vs $0.751$ overall) --- the opposite of what an overfitted prior set would show. The
selection traces make the mechanism explicit: the search chose a different cue set on each fold
and never recovered the paper's four, preferring \texttt{n\_answered} and engagement-rate
features on phd2019 and springer2017 and \texttt{min\_len} with optional-field coverage on
osmi\_mh, and it selected mean-$z$ rather than max-$z$ every time. Its development scores were
high ($0.802$, $0.887$, $0.765$) and fell by $0.07$--$0.24$ once frozen and transferred, which
is the signature of fitting to the development surveys rather than to a population-level
regularity. Per-fold variance is also large --- the frozen configuration ranges from $0.473$ to
$0.889$ across individual cells --- so we report the fold means rather than any single cell.

\begin{table}[t]
\centering\small
\caption{Leave-one-survey-out validation, mean AUROC over the four agentic modes. \emph{Fitted (blind)} selects cues, directions and aggregator by greedy search on the other two surveys and is then frozen; \emph{SPABD} applies the same four cues and sign priors everywhere. Baselines are the best existing detector per survey (Tab.~\ref{tab:main_results}).}
\label{tab:loso}
\begin{tabular}{lccc}
\toprule
Held out & Fitted (blind) & SPABD & Best existing \\
\midrule
phd2019      & 0.729 & \textbf{0.753} & 0.61 \\
springer2017 & 0.651 & \textbf{0.687} & 0.61 \\
osmi\_mh     & 0.624 & \textbf{0.813} & 0.59 \\
\midrule
Mean         & 0.668 & \textbf{0.751} & 0.61 \\
\bottomrule
\end{tabular}
\end{table}

\section{Behavioural Evidence for Agent\_Fill}
\label{sec:appendix_b}

\begin{figure*}[t]
\centering
\includegraphics[width=\textwidth]{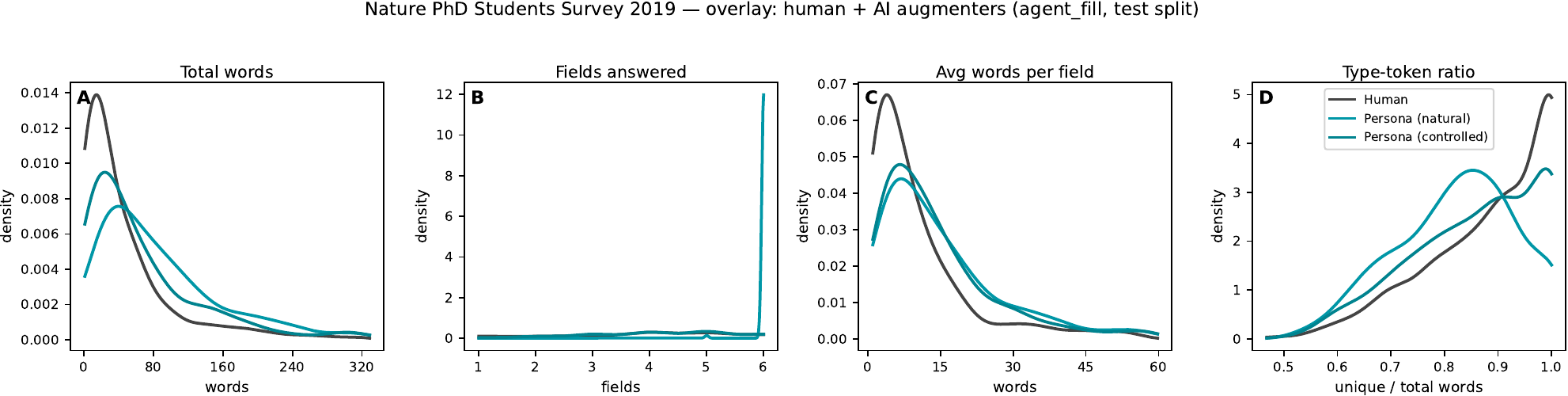}\\[0.5em]
\includegraphics[width=\textwidth]{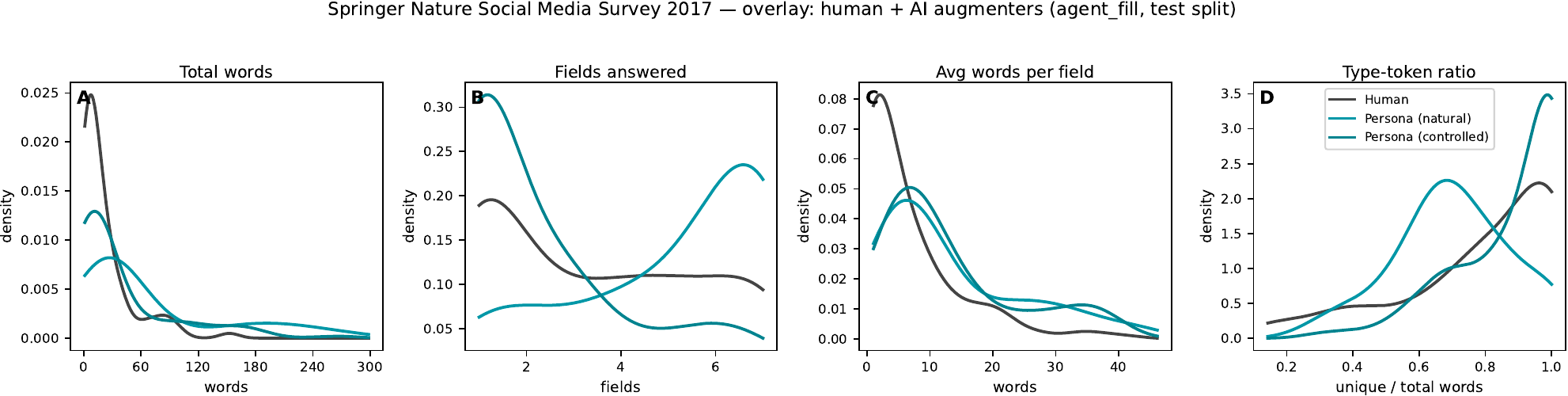}\\[0.5em]
\includegraphics[width=\textwidth]{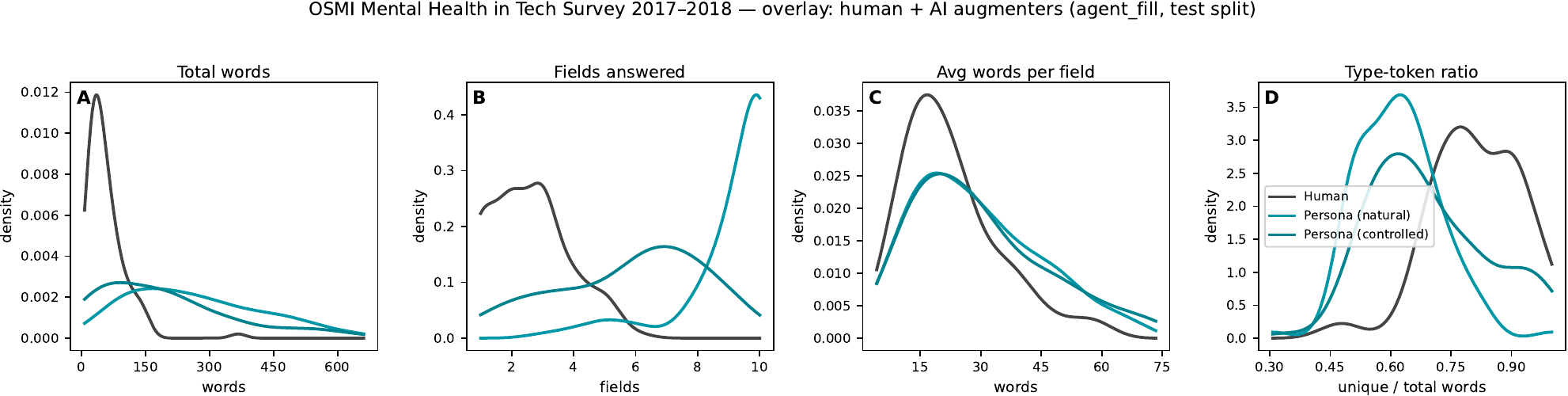}
\caption{Per-respondent behavioural-stats comparison (human vs persona-grounded AI under four prompting modes) across phd2019 (top), springer2017 (middle), and osmi\_mh (bottom). Four panels each: total words, fields answered, average words per field, type--token ratio (TTR).}
\label{fig:appendix_b1}
\end{figure*}

\begin{figure*}[t]
\centering
\includegraphics[width=0.8\textwidth]{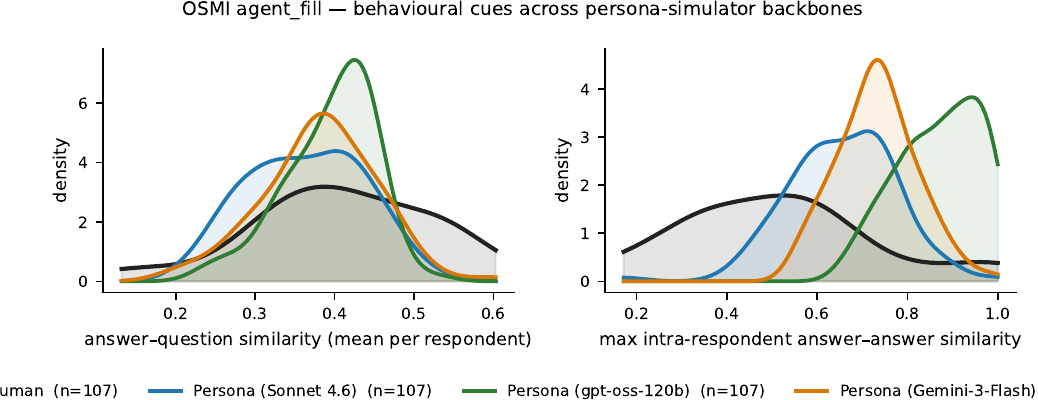}
\caption{Cross-LLM behavioural cues on osmi\_mh: \emph{mean\_qa\_sim} (left) and \emph{max\_aa\_sim} (right) KDEs for human respondents and three persona-simulator backbones (Sonnet, gpt-oss-120b, Gemini-3-Flash). Sonnet and Gemini sit just to the right of the human distribution on \emph{mean\_qa\_sim}, supporting the LLM-family-independent on-topic-ness cue; gpt-oss-120b exhibits a sharp \emph{max\_aa\_sim} shift due to intra-respondent answer reuse.}
\label{fig:appendix_b2}
\end{figure*}

This appendix substantiates the behavioural-divergence claims of Sec.~\ref{sec:analysis}. Tab.~\ref{tab:appendix-b1} reports per-respondent cue distributions (median [IQR]) across the four agent\_fill prompting modes and the three surveys. Fig.~\ref{fig:appendix_b1} visualises the same comparison via four respondent-level statistics (total words, fields answered, average words per field, TTR) on each dataset. Fig.~\ref{fig:appendix_b2} demonstrates that the on-topic-ness and answer-reuse cues are LLM-family-independent across three persona-simulator backbones.

\begin{table*}[t]
\centering\small
\caption{Per-respondent behavioural cues: median [IQR] for human respondents and AI completions under four prompting modes (natural plus three adversarial: controlled, length-restrictive, combined), across three surveys. AI rows use persona-grounded completion with the Sonnet 4.6 backbone.}
\label{tab:appendix-b1}
\begin{tabular}{ll ccc}
\toprule
Cue & Group & PhD & Springer & OSMI \\
\midrule
min\_len & Human & 2 [1,2] & 1 [1,3] & 10 [4,22] \\
 & Natural & 6 [2,13] & 8 [3,14] & 20 [10,36] \\
 & Controlled &6 [3,12] & 9 [5,17] & 20 [13,32] \\
 & Length-restrictive &2 [2,5] & 1 [1,3] & 9 [2,14] \\
 & Combined & 3 [2,6] & 3 [1,6] & 11 [7,19] \\
\midrule
length\_cv & Human & 0.67 [0.42,0.93] & 0.32 [0.00,0.65] & 0.42 [0.01,0.64] \\
 & Natural & 0.33 [0.24,0.43] & 0.19 [0.11,0.28] & 0.18 [0.13,0.25] \\
 & Controlled &0.30 [0.19,0.42] & 0.00 [0.00,0.17] & 0.16 [0.10,0.22] \\
 & Length-restrictive &0.48 [0.38,0.59] & 0.41 [0.30,0.52] & 0.33 [0.26,0.40] \\
 & Combined & 0.45 [0.32,0.57] & 0.25 [0.00,0.41] & 0.30 [0.19,0.38] \\
\midrule
mean\_qa\_sim & Human & 0.23 [0.18,0.28] & 0.17 [0.12,0.27] & 0.42 [0.34,0.50] \\
 & Natural & 0.26 [0.21,0.31] & 0.38 [0.28,0.48] & 0.36 [0.31,0.42] \\
 & Controlled &0.28 [0.22,0.33] & 0.37 [0.27,0.48] & 0.38 [0.32,0.43] \\
 & Length-restrictive &0.25 [0.21,0.30] & 0.28 [0.20,0.35] & 0.33 [0.27,0.38] \\
 & Combined & 0.27 [0.22,0.32] & 0.28 [0.19,0.37] & 0.35 [0.32,0.41] \\
\midrule
n\_answered & Human & 4 [3,5] & 3 [1,5] & 2 [2,3] \\
 & Natural & 6 [6,6] & 6 [4,7] & 10 [9,10] \\
 & Controlled &4 [3,5] & 2 [1,3] & 6 [4,8] \\
 & Length-restrictive &6 [6,6] & 7 [7,7] & 10 [10,10] \\
 & Combined & 4 [3,5] & 2 [1,3] & 7 [4,8] \\
\bottomrule
\end{tabular}
\end{table*}

\section{Qualitative Examples}
\label{sec:qualitative}

We reproduce one representative respondent per survey, with every question they answered shown across all seven variants: the original \emph{Human}, an \emph{LLM Revision} of the human draft, a \emph{Full Generation} from the question alone, and four \emph{Agentic} completions (natural / controlled / length-restrictive / combined). LLM Revision and Full Generation are per-field pipelines that sample 50\% of a respondent's answered fields i.i.d.; where the chosen respondent's field was not sampled we show a representative AI output for that same question from another respondent, marked \textit{\footnotesize(repr.)}. An em-dash marks variants where neither path is available. At the field level the AI variants are visually indistinguishable from one another; the AI signal lives at the respondent level (Sec.~\ref{sec:analysis}, App.~\ref{sec:appendix_b}). Tabs.~\ref{tab:qualitative-phd-a}--\ref{tab:qualitative-osmi-b}.

\begin{table}[t]
\centering\small
\caption{Detection under partial adoption: share of AI respondents in the screened pool varied from 5\% to 50\%, mean over the 12 agentic cells and 20 pool subsamples, $K{=}5$ in-survey reference throughout. Prec@10 is the precision of the ten highest-ranked respondents, i.e.\ a fixed review budget.}
\label{tab:prevalence}
\begin{tabular}{lcccc}
\toprule
AI share of pool & 5\% & 10\% & 25\% & 50\% \\
\midrule
AUROC        & 0.743 & 0.742 & 0.745 & 0.744 \\
Prec@10      & 0.288 & 0.434 & 0.582 & 0.671 \\
Prec@$n_{\mathrm{AI}}$ & 0.251 & 0.300 & 0.486 & 0.628 \\
\bottomrule
\end{tabular}
\end{table}


\definecolor{qhdr}{gray}{0.88}

\newcommand{\qrow}[1]{\rowcolor{qhdr}\multicolumn{2}{@{}p{15.6cm}@{}}{\textit{#1}} \\}
\newcommand{\qaddspace}{\addlinespace[0.2em]}
\newcommand{\repr}{\textit{\footnotesize\,(repr.)}}

\begin{table*}[t]
\centering\small
\renewcommand{\arraystretch}{1.1}
\begin{tabular}{@{}p{2.3cm} p{13.3cm}@{}}
\toprule
\textbf{Variant} & \textbf{Response} \\
\midrule
\qrow{Q1: Overall, how would you describe the academic system, based on your PhD experience so far?}
Human                 & The system needs a complete overhaul. \\
LLM Revision          & A competitive yet vibrant environment. \\
Full Generation       & Honestly, it's a bit of a love-hate thing -- lots of pressure and red tape, but also decent mentorship and some cool collaborations when you find the right people. \\
Agentic (natural)     & Outdated and honestly not in tune with global standards -- supervision is lacking too. \\
Agentic (controlled)  & Idealistic but disconnected from industry, I think. there's real freedom to explore ideas which I genuinely enjoy, but the workload expectations are pretty unrealistic. \\
Agentic (length-rest.)& Kinda disconnected from industry, which bothers me. And there's alot of unpaid work wrapped up as ``opportunity'' -- it gets old pretty fast. \\
Agentic (combined)    & Its quite rigid and honestly pretty disconnected from industry. The RA work just piles up and your actual research suffers. \\
\qaddspace
\qrow{Q2: Is there anything else not mentioned that has concerned you since you started your PhD?}
Human                 & Inadequate or lack of expert supervisors/mentors to guide in the doctoral study and out of date curriculum which is not in tune with global standards as well as too much work load giving little room for research. \\
LLM Revision          & --- \\
Full Generation       & --- \\
Agentic (natural)     & Workload honestly, and the resources here are very limited. \\
Agentic (controlled)  & The RA role honestly takes up way more time than it should, and it directly cuts into my research hours. \\
Agentic (length-rest.)& The RA work takes up way more time than it should, its hard to focus on my own research rn. \\
Agentic (combined)    & The RA workload honestly. It eats into research time more than I expected. \\
\qaddspace
\qrow{Q3: With hindsight, what one thing do you wish you'd known when you started your PhD?}
Human                 & That it is better to study outside my environment. \\
LLM Revision          & Manage my time more wisely and act on it. \\
Full Generation       & I wish I'd known that managing expectations and setting realistic milestones is crucial -- lots of people dive into research without a clear plan. \\
Agentic (natural)     & That doing a PhD abroad would have been a much better option for me. \\
Agentic (controlled)  & How much the RA role would eat into my actual research time. would've pushed back before agreeing to anything. \\
Agentic (length-rest.)& That RA commitments can quietly swallow your entire schedule if you dont set boundaries early on. \\
Agentic (combined)    & Time management is harder than you expect, especially when RA duties keep creeping in. \\
\bottomrule
\end{tabular}
\caption{Qualitative example~1 (Q1--Q3) --- \textbf{Nature PhD Students Survey 2019}, respondent \texttt{SNAP 2015 SHIFT} (persona \texttt{persona\_476}). Em-dashes mark variants where no response was produced; \repr{} marks Revision/Full-Generation outputs taken from another respondent (50\% per-field sampling in these two pipelines).}
\label{tab:qualitative-phd-a}
\end{table*}

\begin{table*}[t]
\centering\small
\renewcommand{\arraystretch}{1.1}
\begin{tabular}{@{}p{2.3cm} p{13.3cm}@{}}
\toprule
\textbf{Variant} & \textbf{Response} \\
\midrule
\qrow{Q4: What type of career are you interested in pursuing after your degree?}
Human                 & Academia. \\
LLM Revision          & University academic staff. \\
Full Generation       & I'm aiming for a data-science/AI role in the tech industry, preferably as a machine-learning engineer. \\
Agentic (natural)     & Academic career, ideally in a university setting. \\
Agentic (controlled)  & R\&D, something that bridges academia and industry. I dont wanna stay purely in academia tbh. \\
Agentic (length-rest.)& R\&D, something that bridges academia and industry. \\
Agentic (combined)    & R\&D, bridging academia and industry. \\
\qaddspace
\qrow{Q5: Any more comments?}
Human                 & No. \\
LLM Revision          & --- \\
Full Generation       & --- \\
Agentic (natural)     & Nothing much more to add, it's been a frustrating experience overall. \\
Agentic (controlled)  & Just that the RA workload issue is something universities need to take seriously. \\
Agentic (length-rest.)& Not really, I think I covered the main stuff. \\
Agentic (combined)    & Not really. Just hope the findings are useful for improving PhD structures. \\
\qaddspace
\qrow{Q6: Overall, what do you enjoy most about life as a PhD student?}
Human                 & Knowing I will have a chance to use my skills in a non-research science job. \\
LLM Revision          & University/Academic Environment. \\
Full Generation       & I love the intellectual freedom -- getting to dive deep into topics I'm really passionate about. \\
Agentic (natural)     & Not much honestly, maybe the independence sometimes. \\
Agentic (controlled)  & The creative side of it, honestly. getting to dig into a problem that actually interests me. \\
Agentic (length-rest.)& The freedom to explore ideas. And honestly the creative side of research. \\
Agentic (combined)    & The creative side of research, honestly. Coming up with ideas that could actually matter in industry. \\
\bottomrule
\end{tabular}
\caption{Qualitative example~1 (Q4--Q6), continued.}
\label{tab:qualitative-phd-b}
\end{table*}

\begin{table*}[t]
\centering\small
\renewcommand{\arraystretch}{1.1}
\begin{tabular}{@{}p{2.3cm} p{13.3cm}@{}}
\toprule
\textbf{Variant} & \textbf{Response} \\
\midrule
\qrow{Q1: Are there any other tasks you use social media/SCN for, in relation to your work?}
Human                 & to stay up-to-date on the latest advances in my areas of research. \\
LLM Revision          & Fortunately, I can find clips and movies related to my research topic on social media.\repr \\
Full Generation       & I also use it for recruiting candidates, posting conference updates, and staying on top of what competitors are doing.\repr \\
Agentic (natural)     & Monitoring my digital presence and occasionally engaging with science communication to non-specialist audiences, which I find aligns well with Open Science goals. \\
Agentic (controlled)  & Monitoring field discussions and curating content for science communication to broader audiences. Also building a consistent professional online presence -- honestly quite useful for Open Science visibility. \\
Agentic (length-rest.)& Keeping up with open science discussions and policy. \\
Agentic (combined)    & Building professional visibility and staying updated on field developments. \\
\qaddspace
\qrow{Q2: Comments on the usefulness of SCN for sharing professional content?}
Human                 & (1) building one's own brand, value, reputation -- building one's own professional E-persona; I consider this a very important aspect of everyone's career in this digital age. (2) supporting / promoting Open Science -- making scientific information available to a wide audience. \\
LLM Revision          & I don't find social media particularly useful for this, since it still has to earn credibility---except in a few cases, of course.\repr \\
Full Generation       & Social media is useful for getting quick updates and connecting with peers, but the sheer volume of content can make it hard to find really relevant professional info.\repr \\
Agentic (natural)     & Sharing content through SCNs has been pretty effective for me -- preprints, conference talks, and outputs that might otherwise sit behind paywalls reach a broader audience. \\
Agentic (controlled)  & Sharing papers and preprints through SCNs has been effective for reaching audiences beyond my immediate network. it's also a good way to establish visibility for Open Science outputs. \\
Agentic (length-rest.)& Useful for reaching audiences beyond the journal. \\
Agentic (combined)    & Useful for reaching audiences beyond traditional publishing channels, especially for open access content. \\
\qaddspace
\qrow{Q3: Anything that publishers could provide or improve to help your professional use of SCN?}
Human                 & perhaps present more information on science \& policy, funding opportunities via social media channels (more emphasis on sharing news \& views, commentaries, blog articles related to these fields). \\
LLM Revision          & Pop-ups should be blocked. Instead, send an email---perhaps offering a free reading trial---to encourage users.\repr \\
Full Generation       & A unified scheduling tool with clearer analytics and a way to filter out irrelevant content would be super helpful.\repr \\
Agentic (natural)     & Better integration between journal platforms and SCNs would help -- something like automatic DOI-linked sharing options at point of publication. Open access by default would also do more than any platform feature. \\
Agentic (controlled)  & Better integration between journal platforms and SCNs -- making it easier to share newly published work with a single click and track engagement metrics. \\
Agentic (length-rest.)& Better integration between journal platforms and SCNs, and clearer guidelines on sharing preprints/accepted manuscripts. \\
Agentic (combined)    & Better integration between journal platforms and SCNs for sharing publications directly. \\
\qaddspace
\qrow{Q4: Any further comments on your professional use of social media/SCN?}
Human                 & no. \\
LLM Revision          & No. \\
Full Generation       & I use it mainly for networking and staying up-to-date -- LinkedIn for industry news and Twitter for quick updates.\repr \\
Agentic (natural)     & The main shift I've noticed is that maintaining a professional digital presence feels less optional than it did -- it's increasingly part of how research impact gets measured and recognised, whether we like it or not. \\
Agentic (controlled)  & Publishers and platforms could do more to recognise SCN activity as a legitimate form of scholarly output -- engagement metrics and public communication efforts are still largely invisible in academic evaluation. \\
Agentic (length-rest.)& More open access by default would help a lot --- it makes sharing research far easier. \\
Agentic (combined)    & No strong further comments --- Open Science advocacy is where I find these tools most valuable. \\
\bottomrule
\end{tabular}
\caption{Qualitative example~2 --- \textbf{Springer Nature Social Media Survey 2017}, respondent \texttt{R\_2cvJRdYeEQWgy0z} (persona \texttt{persona\_042}). Four questions, seven variants each.}
\label{tab:qualitative-springer}
\end{table*}

\begin{table*}[t]
\centering\small
\renewcommand{\arraystretch}{1.1}
\begin{tabular}{@{}p{2.3cm} p{13.3cm}@{}}
\toprule
\textbf{Variant} & \textbf{Response} \\
\midrule
\qrow{Q1: Describe the conversation you had with your employer about your mental health.}
Human                 & I disclosed my OCD/anxiety issues, and I felt safe and heard. Next steps included allowing for any accommodations I needed, including working from home more often. \\
LLM Revision          & I have severe anxiety and ADD, and I take medication for both. Occasionally this can affect my work performance, but I make a concerted effort to prevent any impact.\repr \\
Full Generation       & I told my manager I was feeling burnt out and pretty anxious. They were supportive, cut back my workload, gave me flexible hours, and pointed me to the EAP.\repr \\
Agentic (natural)     & Told my manager during a rough stretch. they were pretty understanding, made some small adjustments -- honestly better than I expected. \\
Agentic (controlled)  & Only disclosed to folks I was pretty confident would be supportive, and honestly it went better than I expected each time. \\
Agentic (length-rest.)& Only brought it up with my manager -- they were pretty understanding. we worked out some flexibility around deadlines. \\
Agentic (combined)    & Kept it fairly high-level with my manager -- mentioned I was dealing with some anxiety and might need flexibility. \\
\qaddspace
\qrow{Q2: Describe the conversation with coworkers about your mental health.}
Human                 & It's always been well received, I've felt supported. However, I've only disclosed to folks who I anticipated would be supportive. \\
LLM Revision          & I have struggled with depression and panic attacks.\repr \\
Full Generation       & I mentioned to a couple of teammates that I've been feeling pretty anxious lately. One suggested grabbing coffee; another just said, ``Take care of yourself.''\repr \\
Agentic (natural)     & Only disclosed to folks I was pretty sure would be okay with it. they were supportive, no drama. \\
Agentic (controlled)  & Similar story -- only brought it up with a couple people I trusted. they were pretty understanding, no awkwardness about it. \\
Agentic (length-rest.)& Only told a couple folks I trusted. they were supportive, no weird reactions or anything. \\
Agentic (combined)    & Only disclosed to folks who I anticipated would be supportive. those conversations went fine. \\
\qaddspace
\qrow{Q3: Describe the conversation your coworker had with you about their mental health.}
Human                 & It was mentioned that this colleague was suffering from depression, and I was asked for advice on how best to support. \\
LLM Revision          & I was told that this colleague was dealing with depression, and I was asked for advice on the best ways to support them. \\
Full Generation       & They said they've been feeling pretty overwhelmed and anxious lately, especially with the workload. They mentioned they're seeing a therapist and have started medication, and asked if I could be flexible with deadlines. \\
Agentic (natural)     & A colleague came to me during a hard time -- kept it pretty vague but I could tell it meant a lot to just have someone listen. \\
Agentic (controlled)  & A coworker opened up to me about burnout and some anxiety. I just tried to listen, not make it weird. \\
Agentic (length-rest.)& They mentioned they were struggling with burnout, kind of just venting. I just listened, didn't make it a big deal. \\
Agentic (combined)    & A colleague mentioned they were struggling with burnout. I just listened, didn't make it weird. \\
\bottomrule
\end{tabular}
\caption{Qualitative example~3 (Q1--Q3) --- \textbf{OSMI Mental Health in Tech 2017--18}, respondent \texttt{osmi\_2018\_0200} (persona \texttt{persona\_035}).}
\label{tab:qualitative-osmi-a}
\end{table*}

\begin{table*}[t]
\centering\small
\renewcommand{\arraystretch}{1.1}
\begin{tabular}{@{}p{2.3cm} p{13.3cm}@{}}
\toprule
\textbf{Variant} & \textbf{Response} \\
\midrule
\qrow{Q4: Describe the conversation with your previous employer about your mental health.}
Human                 & I disclosed my OCD/anxiety to my whole team. It was awkward, but they were ultimately caring and supportive. \\
LLM Revision          & I disclosed my OCD and anxiety to my entire team. Although it felt awkward at first, they responded with care and support. \\
Full Generation       & I sat down with my manager and HR after noticing I was getting overwhelmed and told them I was struggling with anxiety and burnout. They were pretty understanding -- my manager cut back my overtime and let me work from home a couple days a week. \\
Agentic (natural)     & Pretty similar experience -- disclosed to my direct manager when things got tough, they were receptive. \\
Agentic (controlled)  & Only disclosed to one manager there, someone I had a good read on. they were fine about it, offered some flexibility which helped. \\
Agentic (length-rest.)& Similar experience -- told my manager at my previous job when I needed accommodations. \\
Agentic (combined)    & Similar experience -- mentioned I had OCD and anxiety, kept it brief. My previous manager was supportive, offered some schedule flexibility. \\
\qaddspace
\qrow{Q5: Describe the conversation with your previous coworkers about your mental health.}
Human                 & Again, I only disclosed when I felt safe enough to do so. I felt supported and heard by those folks. \\
LLM Revision          & I told a few coworkers that I was seeing a therapist, both to help break down stigma and to encourage them to seek mental-health support.\repr \\
Full Generation       & I was pretty open about feeling burnt out and anxious. A few teammates just said ``take care of yourself, we've all been there.''\repr \\
Agentic (natural)     & Pretty similar to my current job honestly -- only told a few people I trusted. \\
Agentic (controlled)  & Similar to current job -- only disclosed to folks who I anticipated would be supportive. \\
Agentic (length-rest.)& Kept it pretty limited, only disclosed to folks who I anticipated would be supportive. \\
Agentic (combined)    & Only told one or two people. reactions were fine, they didn't make a thing of it. \\
\qaddspace
\qrow{Q6: What could the industry/employers do to improve mental health support?}
Human                 & Obsessive-Compulsive Disorder. \\
LLM Revision          & I have been diagnosed with Obsessive-Compulsive Disorder (OCD). \\
Full Generation       & Employers could make mental-health resources a regular part of the work culture -- offering free counseling, flexible schedules, and clear policies that reduce stigma. \\
Agentic (natural)     & More training for managers on how to actually respond -- not just policies on paper. \\
Agentic (controlled)  & More openness from leadership would go a long way -- if managers talk about it, it normalizes things for everyone. \\
Agentic (length-rest.)& More openness at the leadership level, I think. concrete stuff like flexible time off and EAP that people actually know about. \\
Agentic (combined)    & More openness at the leadership level would help -- if managers talk about it, it normalizes things. \\
\bottomrule
\end{tabular}
\caption{Qualitative example~3 (Q4--Q6), continued.}
\label{tab:qualitative-osmi-b}
\end{table*}

\section{Deployment Metrics}
\label{sec:deployment}

The main paper reports AUROC because it is the standard benchmark metric in AI-text detection. For deployment, practitioners care more about operating-point behaviour: how many real respondents are wrongly flagged for each AI submission caught, and how well the detector ranks AI cases when they are the rare class. Tab.~\ref{tab:deployment-metrics} in the main text reports AUPR (rank quality under class imbalance) and TPR at low FPR (the catch rate at deployment-relevant false-alarm budgets), averaged across the 12 agent\_fill cells; this appendix gives the metric definitions.

\paragraph{Definitions.} \textbf{AUPR} is the area under the precision--recall curve; values range in $[0,1]$ with the chance level equal to the AI prevalence. \textbf{TPR@FPR=$x$\%} is the fraction of AI respondents caught at the decision threshold where at most $x\%$ of humans are wrongly flagged --- the natural metric when each false positive corresponds to a wrongfully rejected respondent.

\paragraph{Partial adoption.} The operating points above assume a balanced pool. In
practice only some respondents use AI, so we vary the AI share of the screened pool from
$5\%$ to $50\%$, holding the four cues and the $K{=}5$ in-survey reference fixed and averaging
over repeated pool draws (Tab.~\ref{tab:prevalence}). Ranking quality is essentially
prevalence-invariant --- AUROC stays within $0.003$ of $0.744$ across the whole range ---
because each respondent is scored against a human reference rather than against the rest of the
pool. What changes is the yield of a fixed review budget: the precision of the ten
highest-ranked respondents rises from $0.288$ at $5\%$ adoption to $0.671$ at $50\%$. A
practitioner screening a lightly affected survey should therefore expect to inspect several
genuine respondents for each AI submission found, and that ratio improves as adoption grows.

\section{Demographic-Bias Analysis}
\label{sec:bias}

\citet{liang2023gpt} show that perplexity-based detectors disproportionately flag
non-native English writers as AI-generated, on the basis of TOEFL essays compared
against essays by native students. That setting differs from ours in both the
conditions of production (a timed high-stakes exam versus a voluntary survey) and
the response characteristics (long-form essays versus short, fragmented answers
spread over multiple fields). SPABD is also not a perplexity detector and reads no
fluency or vocabulary signal; of the properties implicated by \citet{liang2023gpt},
only length enters our cue set. We nonetheless test directly whether the reported
bias appears in survey completion.

For each survey we identify non-native English respondents by interface language or
country of residence and compare them against native respondents on the SPABD cues
and on the resulting flag rates. Answer-length variance is not significantly more
rigid for non-native respondents on any survey (phd2019 $p = 0.21$, springer2017
$p = 0.46$, osmi\_mh $p = 0.22$), so the \texttt{length\_cv} cue does not encode
proficiency. On phd2019, the largest survey ($n = 127$ non-native and $476$ native
respondents with a usable proxy), non-native answers are \emph{shorter} and
\emph{less} on-topic than native ones --- the opposite of the AI direction on both
\texttt{min\_len} (effect AUC $0.31$) and \texttt{mean\_qa\_sim} ($0.09$) --- and are
correspondingly flagged less often, $0.6\%$ versus $2.4\%$ at a global $5\%$
operating point. springer2017 shows the same direction ($3.0\%$ versus $7.5\%$, with
$n = 101$ and $45$). One cue runs the other way: on phd2019 non-native respondents
answer slightly more fields (\texttt{n\_answered}, effect AUC $0.59$, $p = 0.002$),
which is the AI direction, but it is outweighed by the other two. On osmi\_mh
non-native respondents show a higher flag rate ($14.2\%$), but with only $n = 10$
matched respondents this is within sampling noise. We also probed gender and age band
on osmi\_mh and geographic region on phd2019 and springer2017: no cue differs
significantly by gender or age band, while regional differences track the same length
and on-topic-ness effects as the language proxy.

We report this as evidence against the specific bias documented by
\citet{liang2023gpt} in this setting, not as a general fairness guarantee. Proficiency
proxies derived from interface language and country are coarse, the osmi\_mh subgroup
is too small to support a conclusion, and the subgroup axes available to us are limited
to what the source surveys recorded. As
stated in the Ethical Considerations, flagged responses should receive human review
rather than automatic rejection.

\end{document}